\documentclass[journal,11pt,doublespace,twocolumn]{IEEEtran}

\usepackage{amsmath}
\usepackage{graphicx,psfrag,epsf}
\usepackage{enumerate}
\usepackage{url} 
\usepackage{url} 
\usepackage{amssymb,amsfonts,mathtools}
\usepackage[lofdepth,lotdepth]{subfig}
\usepackage{bm}
\usepackage{color, colortbl}
\usepackage{algorithm}
\usepackage{algorithmic}
\usepackage{balance,cite}
\usepackage{multirow}
\usepackage{cuted}

\usepackage{amsthm}
\makeatletter
\def\th@plain{%
  \thm@notefont{}
  \itshape 
}
\def\th@definition{%
  \thm@notefont{}
  \normalfont 
}
\makeatother

\DeclareMathOperator*{\argmin}{arg\,min}
\DeclareMathOperator*{\argmax}{arg\,max}
\newcommand{\rom}[1]{\uppercase\expandafter{\romannumeral #1\relax}}
\DeclarePairedDelimiterX{\norm}[1]{\lVert}{\rVert}{#1}
\DeclarePairedDelimiterX{\bnorm}[1]{\biggl\lVert}{\biggr\rVert}{#1}
\DeclarePairedDelimiterX{\abs}[1]{\lvert}{\rvert}{#1}

\renewcommand{\baselinestretch}{1} 

\newtheorem{definition}{Definition}

\newtheorem{example}{Example}

\def\T{{ \mathrm{\scriptscriptstyle T} }}
\def\P{{ \mathrm{pr} }}
\def\v{{\varepsilon}}

\def\E{E_{*}}

\def\P{P_*}
\def\de{\overset{\Delta}{=}}

\def\limp{\rightarrow_{p}}
\def\limd{\rightarrow_{d}}
\def\moind{\mathbb{M}}
\def\morep{\mathcal{M}_m}
\def\mo{m}
\def\H{\mathcal{H}_m}
\def\Mo{\{\morep\}_{m \in \moind}}

\def\th{\theta}

\def\loss{\mathcal{L}} 
\def\p{d_n} 
\def\d{d_m}

\def\LM{d_n}
\def\AIC{\textsc{aic}}
\def\AICc{\textsc{aic}\textrm{c}}
\def\BC{\textsc{bc}}
\def\HQ{\textsc{hq}}
\def\BIC{\textsc{bic}}

\def\PLS{\textsc{pls}}
\def\DIC{\textsc{dic}}

\def\GIC{\textsc{gic}_{\lambda_n}}
\def\PI{\textsc{pi}_n}

\def\CO{}

\def\coco{}

\begin{document}

\title{Model Selection Techniques \\ ---An Overview }

\author{Jie~Ding,~
        Vahid Tarokh,~
        and~Yuhong Yang
\thanks{This research was funded in part by the Defense Advanced Research Projects Agency (DARPA) under grant number W911NF-18-1-0134. } 
\thanks{J.~Ding and Y.~Yang are with the School of Statistics, University of Minnesota, Minneapolis, Minnesota 55455, United States.
V. Tarokh is with the Department of Electrical and Computer Engineering, Duke University, Durham, North Carolina 27708, United States.
}
\thanks{
Digital Object Identifier 10.1109/MSP.2018.2867638 \copyright2018 IEEE
}
}

\markboth{IEEE Signal Processing Magazine
}
{Shell \MakeLowercase{\textit{et al.}}: IEEE Transactions on Signal Processing}

\maketitle

\begin{abstract}

In the era of ``big data'', analysts usually explore various statistical models or machine learning methods for observed data in order to facilitate scientific discoveries or gain predictive power. Whatever data and fitting procedures are employed, a crucial step is to select the most appropriate model or method from a set of candidates.  Model selection is a key ingredient in data analysis for reliable and reproducible statistical inference or prediction, and thus  central to scientific studies in fields such as ecology, economics, engineering, finance, political science, biology, and epidemiology. 
There has been a long history of model selection techniques that arise from researches in statistics, information theory, and signal processing. A considerable number of methods have been proposed, following different philosophies and exhibiting varying performances. The purpose of this article is to bring a comprehensive overview of them, in terms of their motivation, large sample performance, and applicability. 
We provide integrated and practically relevant discussions on theoretical properties of state-of-the-art model selection approaches.
We also share our thoughts on some controversial views on the practice of model selection.
  
\end{abstract}



\section{Why Model Selection} \label{sec:why}

Vast development in hardware storage, precision instrument manufacture, economic globalization, etc. have generated huge volumes of data that can be analyzed to extract useful information. 
Typical  statistical inference or machine learning procedures learn from and make predictions on data by fitting parametric or nonparametric
models (in a broad sense). However, there exists no model that is universally suitable for any data and goal.   
An improper choice of model or method can lead to purely noisy ``discoveries'', severely misleading conclusions, or disappointing predictive performances. 
Therefore, a crucial step in a typical data analysis is to consider a set of candidate models (referred to as the \textit{model class}),  and then select the most appropriate one. 
In other words, \textit{model selection} is the task of selecting a statistical model  from a model class, given a set of data. 
For example, we may be interested in the selection of 
\begin{itemize}
  \item variables for linear regression,
  \item basis terms such as polynomials, splines, or wavelets in function estimation,
  \item order of an autoregressive process, 
  \item number of components in a mixture model, 
  \item most appropriate parametric family among a number of alternatives, 
  \item number of change points in time series models,    
  \item number of neurons and layers in neural networks,
  \item best choice among logistic regression, support vector machine, and neural networks,
  \item best machine learning techniques for solving real-data challenges on an online competition platform.
\end{itemize}

%
There have been many overview papers 
on model selection scattered in the communities of 
signal processing~\cite{stoica2004model}, 
statistics~\cite{kadane2004methods}, 
machine learning~\cite{guyon2003introduction},
epidemiology~\cite{greenland1989modeling},
chemometrics~\cite{andersen2010variable}, 
ecology and evolution~\cite{johnson2004model}. 
Despite the abundant literature on model selection, existing overviews usually focus on derivations, descriptions, or applications of particular model selection principles. In this paper, we aim to provide an integrated understanding of the properties and practical performances of various approaches, by reviewing their theoretical and practical advantages, disadvantages, and relations.
Our overview is characterized by the following highlighted perspectives, which we believe will bring deeper insights for the research community.

First of all, we provide a 
	technical overview of foundational model selection approaches and their relations, by centering around two statistical goals, {\coco namely} prediction and inference (see Section~\ref{sec:criteria}). 
We then introduce rigorous developments in the understanding of two fundamentally representative model selection criteria (see Section~\ref{sec:AICBIC}). 
We also review research developments in the  problem of high-dimensional variable selection, including penalized regression and step-wise variable selection which have been widely used in practice (see Section~\ref{sec:high}). 
Moreover, we review recent developments in modeling procedure selection, which, differently from model selection in the narrow sense of choosing among parametric models, aims to select the better statistical or machine learning procedure (see Section~\ref{sec:selecting}). 
Finally, we address some common misconceptions and controversies about 
	model selection techniques (see Section~\ref{sec:clarify}), and provide a few general recommendations on the application of model selection (in Section~\ref{recomm}). 
For derivations and implementations of some popular information criteria, 
we refer the readers to monographs and review papers such as \cite{burnham2003model,stoica2004model}.


\subsection{Some Basic Concepts}

\textbf{Notation}: We use {\coco $\morep = \{p_{\th_m}: \th_m \in \H \}$} to denote a model (in the formal probabilistic sense), which is a set of probability density functions to describe the data $z_1,\ldots,z_n$. Here, $\H$ is the parameter space associated with $\morep$. 
A model class, $\{\morep \}_{m \in \moind}$, is a collection of models indexed by $m \in \moind$. The number of models (or the cardinality of $\moind$) can be fixed, or depend on the sample size $n$. For each model $\morep$, we denote by $\d$ the dimension of the parameter in model $\morep$. 
{\coco Its log-likelihood function is written as 
$ \th_m \mapsto \ell_{n,m}(\th_m)=\log p_{\th_m}(z_1,\ldots, z_n)$, 
and the maximized log-likelihood value is 
\begin{align}
\ell_{n,m}(\hat{\th}_m) \textrm{ with }\hat{\th}_m= \argmax_{\th_m \in \H}p_{\th_m}(z_1,\ldots,z_n) \label{eq:MLE}
\end{align} 
being the maximum likelihood estimator (MLE) under model $\morep$.
We will write $\ell_{n,m}(\hat{\th}_m)$ as $\hat{\ell}_{n,m}$ for simplicity. }
We use $p_*$ and $E_*$ to denote the true-data generating distribution and expectation with respect to the true data-generating distribution, respectively. 
{\CO In the \textit{parametric} framework, there exists some $m \in \moind$ and some $\th_* \in \H$ such that $p_*$ is exactly $p_{\th_*}$.  In the \textit{nonparametric} framework,  $p_*$ is excluded in the model class. 
We sometimes call a model class $\{\morep \}_{m \in \moind}$ {well-specified} (resp. {mis-specified}) if the data generating process is in a parametric (resp. nonparametric) framework.}
We use $\limp$ and $\limd$ to denote convergence in probability and in distribution (under $p_*$), respectively. We use $\mathcal{N}(\mu, V)$ to denote a Gaussian distribution of mean $\mu$ and covariance $V$, $\chi^2_d$ to denote a chi-squared distribution with $d$ degrees of freedom, and $\norm{\cdot}_2$ to denote the Euclidean norm. The word  ``variable'' is often referred to as the ``covariate'' in a regression setting. 

A typical data analysis can be thought of as consisting of two steps:\\
\textbf{Step 1}: For each candidate model $\morep = \{p_{\th_m}, \th_m \in \H \}$, fit all the observed data to that model by estimating its parameter $\th_m \in \H$; \\
\textbf{Step 2}: Once we have a set of estimated candidate models $p_{\tilde{\th}_m} (m \in \moind)$, select the most appropriate one {\coco for either interpretation or prediction.} 

{\CO We note that not every data analysis and its associated model selection procedure formally rely on probability distributions. Examples of model-free methods are nearest neighbor learning, 
reinforcement learning, 
and expert learning. 
Before we proceed, it is helpful to first introduce the following two concepts. 


{\bf The ``model fitting''}: 
The fitting  procedure (also called parameter estimation) given a certain candidate model $\morep$ is usually achieved by minimizing the following (cumulative) loss:
\begin{align}
\tilde{\th}_m = \argmin_{\th_m \in \H} \sum_{t=1}^{n} s(p_{\th_m}, z_t) \label{eq:obj}.
\end{align}
In the objective (\ref{eq:obj}), each $p_{\th_m}$ represents a distribution for the data, and $s(\cdot, \cdot)$, referred to as the loss function (or scoring function), is used to evaluate the ``goodness of fit'' between a distribution and the observation. 
A commonly used loss function is the logarithmic loss
\begin{align}
	s(p, z_t) = -\log p(z_t)  \label{eq:logscore} ,
\end{align}
the negative logarithm of the distribution of $z_t$.
Then, the objective (\ref{eq:obj}) produces the MLE for a parametric model. 
{\coco For time series data, (\ref{eq:logscore}) is written as $-\log p(z_t \mid z_1,\ldots, z_{t-1})$,
and the quadratic loss 
$s(p, z_t) = \{ z_t - E_p(z_t \mid z_1,\ldots, z_{t-1}) \}^2$ is often used,
where the expectation is taken over the joint distribution $p$ of $z_1,\ldots,z_t$.}

{\CO 

{\bf The ``best model''}: 
Let $\hat{p}_m = p_{\tilde{\th}_m}$ denote the estimated distribution under model $\morep$. The predictive performance can be assessed via the out-sample prediction loss, defined as 
\begin{align}
	E_*(s(\hat{p}_m, Z)) = \int  s(\hat{p}_m(z), z) p_*(z) dz \label{eq:outsample}
\end{align}
where $Z$ is independent with and identically distributed as the data used to obtain $\hat{p}_m$. Here, $Z$ does not have the subscript $t$ as it is the out-sample data used to evaluate the predictive performance.  There can be a number of variations to this in terms of the prediction loss function~\cite{parry2012proper} and time dependency.  
In view of this definition, the best model can be naturally defined as the candidate model with the smallest out-sample prediction loss, i.e., 
$$\hat{m}_{0} = \argmin_{m \in \moind} E_*(s(\hat{p}_m, Z)).
$$ 
In other words, $\mathcal{M}_{\hat{m}_{0}}$ is the model whose predictive power is the best offered by the candidate models. 
We note that the ``best'' is in the scope of the available data,  the class of models, and the  loss function. 

In a parametric framework, typically \textit{the true data-generating model, if not too complicated, is the best model}. In this vein, if the true density function $p_*$ belongs to some model $\morep$, or equivalently $p_* = p_{\th_*}$ for some $\th_* \in \H$ and $m \in \moind$, then we seek to select  such $\morep$ (from $\Mo$) with probability going to one as the sample size increases, which is called consistency in model selection. In addition, the MLE of $p_{\th_m}$ for $\th_m \in \H$ is known to attain Cramer-Rao lower bound asymptotically. In a nonparametric framework, the best model depends on the sample size--typically the larger the sample size, the larger the dimension of the best model since more observations can help reveal weak variables (whose effects are relatively small) that are out-of-reach at a small sample size. As a result, the selected model is sensitive to the sample size, and selection consistency becomes statistically unachievable. We revisit this point in Subsection~\ref{subsec:consistency}. 

We note that the aforementioned equivalence between the best model and the true model may not hold for regression settings where the number of independent variables is large relative to the sample size. Here, even if the true model is included as a candidate, its dimension may be too high to be appropriately identified based on relatively small data. Then the parametric framework becomes practically nonparametric. We will emphasize this point in Subsection~\ref{subsec:illustration}.

\subsection{Goals of data analysis and model selection}

There are two main objectives in learning from data. One is for scientific discovery, understanding of the data generation process, and interpretation of the nature of the data. For example, a scientist may use the data to support a physical model or identify genes that clearly promote early onset of a disease. {\coco Another objective of learning from data  is for prediction, i.e., to quantitatively describe  future observations}. Here the data scientist does not necessarily care about obtaining an accurate probabilistic description of the data. Of course, one may also be interested in both directions. 

In tune with the two different objectives above, model selection can also have two directions: \textit{model selection for inference} and \textit{model selection for prediction}. The first one is intended to identify the best model for the data, which hopefully provides a reliable characterization of the sources of uncertainty for scientific insight and interpretation. And the second is to choose a model as a vehicle to arrive at a model or method that offers top performance. For the former goal, it is crucially important that the selected model is not too sensitive to the sample size. For the latter, however, the selected model may be simply the lucky winner among a few close competitors, yet the predictive performance can still be (nearly) the best possible. If so, the model selection is perfectly fine for the second goal (prediction), but the use of the selected model for insight and interpretation may be severely unreliable and misleading.
   
Associated with the first goal of model selection for \textit{inference} or identifying the best candidate is the following concept of \textit{selection consistency}. 
\begin{definition}\label{def:consis}
	A model selection procedure is consistent if the best model is selected with probability going to one as $n \rightarrow \infty$.
\end{definition}
In the context of variable selection, in practical terms, model selection consistency is intended to mean that the important variables are identified and their statistical significance can be ascertained in a followup study of a similar sample size but the rest of the variables cannot.

In many applications, prediction accuracy is the dominating consideration.
Even when the best model as defined earlier cannot be selected with high probability, other models may provide asymptotically equivalent predictive performance. 
The following \textit{asymptotic efficiency} property demands  that the loss of the selected model or method is asymptotically equivalent to the smallest among all the candidate.
\begin{definition}\label{def:eff}
A model selection procedure is asymptotically efficient if 
\begin{align}
 	\frac{ \min_{\mo \in \moind} \loss_m }{ \loss_{\hat{\mo}} } \limp 1 
 	\qquad \textrm{as} \,\, n \rightarrow \infty, \label{eq:generalEfficiency}	
\end{align}
where $\hat{\mo}$ is the selected model, $\loss_m = E_*(s(\hat{p}_m, Z)) - E_*(s(p_*, Z))$ is the adjusted prediction loss, $\hat{p}_m$ denotes the estimated distribution under model $\mo$.
\end{definition}
The above subtraction of $E_*(s(p_*, Z))$ {\coco allows for better comparison of competing model selection methods.} 
Another property often used to describe  model selection is minimax-rate optimality, which will be elaborated on in Subsection~\ref{subsec:theory}. 
A related but different school of thoughts is the structural risk minimization in the literature of statistical learning theory. In that context, a common practice is to bound the out-sample prediction loss using in-sample loss plus a positive term (e.g. a function of the Vapnik-Chervonenkis (VC) dimension~\cite{vapnik1971uniform} for a classification model). The major difference of {\coco the current setting} compared with that in statistical learning is the (stronger) requirement that the selected model should exhibit prediction loss comparable to the best offered by the candidates. In other words, the positive term plus the in-sample loss should asymptotically approach the true out-sample loss (as sample size goes to infinity).

The goals of \textit{inference} and \textit{prediction} as assessed in terms of asymptotic efficiency of model selection can often be well aligned in a parametric framework, although there exists an unbridgeable conflict when a minimax view is taken to assess the prediction performance. We  will elaborate on this and related issues in Section~\ref{sec:AICBIC}. 

}

In light of all the above discussions, we note that the task of \textit{model selection} is primarily concerned with the selection of $\morep \ (m\in \moind)$, because once $m$ is identified, the model fitting part is straightforward. Thus, the model selection procedure can also be regarded as a joint estimation of both the distribution family ($\morep$) and the parameters in each family ($\th_m \in \H$). 

A model class $\Mo$ is nested if smaller models are always special cases of larger models. 
For a nested model class,  
the model selection is sometimes referred to as the \textit{order selection} problem. The task of model selection in its broad sense can also refer to method (or modeling procedure) selection, which we shall revisit in Section~\ref{sec:selecting}. 

\subsection{An illustration on fitting and the best model} \label{subsec:illustration}

We provide a synthetic experiment to illustrates the general rules that  1) better fitting does not imply better predictive performance, and 2) the predictive performance is optimal at a candidate model that typically depends on both the sample size and the unknown data-generating process.
As a result, an appropriate model selection technique is important to single out the best model for inference and prediction in a strong practically parametric framework, or to strike a good balance between the \textit{goodness of fit} and \textit{model complexity} on the observed data to facilitate optimal prediction in a practically nonparametric framework. 

\begin{example}
\label{example:a}
Suppose that a set of time series data $\{z_t: t=1,\ldots,n\}$ is observed, and we specify an autoregressive (AR) model class with order at most $\LM$. Each model of dimension (or order) $k \ (k = 1,\ldots, \LM)$ is in the form of 
    \begin{align} \label{AR}
        z_t = \sum_{i=1}^k \psi_{k,i} z_{t-i}  + \varepsilon_t,
    \end{align}
referred to as the AR$(k)$, 
where 
$\psi_{k, i} \in \mathbb{R}$ ($i=1,\ldots,k$), $\psi_{k, k} \neq 0$, 
and $\varepsilon_t$'s are independent Gaussian noises with zero mean and variance $\sigma^2$.
Adopting quadratic loss, the parameters $\psi_{k,1}, \ldots, \psi_{k,k}$ can be estimated by the method of least squares. 
When the data-generating model is unknown, one critical problem is the identification of the (unknown) order of the autoregressive model. We need to first estimate parameters with different orders $1,\ldots, \LM$, and then select one of them based on a certain principle. 

\end{example} 

{\CO 

\textit{Experiment}:
In this experiment, we first generate time series data using each of the following three true data-generating processes, with the sample sizes $n=100,500,2000,3000$. We then fit the data using the model class in Example~\ref{example:a}, with maximal order $\LM = 15$. 

1) Parametric framework:
The data are generated in the way described by (\ref{AR}) with true order $k_0=3$, and parameters $\psi_{3,\ell}=0.7^{\ell}$ ($\ell=1,2,3$).  

Suppose that we adopt the quadratic loss in Example~\ref{example:a}. Then we obtain the in-sample prediction loss   
\begin{align*}
       \hat{e}_k= (n-k)^{-1} \sum_{t=k +1}^{n} \biggl (z_t - \sum_{i=1}^k  \hat{\psi}_{k,i} z_{t-i}   \biggr)^2 .
\end{align*}
In Figure~\ref{fig:forSlides_AR3}(a), we plot $\hat{e}_k$ against $k$ for $k=1,\ldots,\LM$, averaged over 50 independent replications. The curve for each sample size $n$ is monotonically decreasing, because larger models fit the same data better.
We compute and plot in Figure~\ref{fig:forSlides_AR3}(b) the out-sample prediction loss in (\ref{eq:outsample}), which is equivalent to  
$
E_*(s(\hat{p}_k, Z_t)) = \E (Z_t - \sum_{i=1}^k \hat{\psi}_{k,i} Z_{t-i})^2 $
in this example. 
The above expectation is taken over the true stationary distribution of an independent process of $Z_t$.\footnote{An alternative definition is based on the same-realization expectation that calculates the loss of the future of an observed 
time series~\cite{ing2005orderselection}.} 
The curves in Figure~\ref{fig:forSlides_AR3}(b) show that the predictive performance is only optimal at the true order. 

Under the quadratic loss, we have $E_*(s(p_*, Z_t)) = \sigma^2$, and the {asymptotic efficiency} (Definition~\ref{def:eff}) requires that 
\begin{align}
&\frac{\min_{k=1,\ldots,d_n} \E (Z_t - \sum_{i=1}^k \hat{\psi}_{k,i} Z_{t-i} )^2 - \sigma^2}{\E (Z_t - \sum_{j=1}^{\hat{k}} \hat{\psi}_{\hat{k},j} Z_{t-j} )^2 - \sigma^2} \label{eq:efficiency}
\end{align}
converges to one in probability. 
In order to describe how the predictive performance of each model deviates from the best possible, we define the \textit{efficiency} of each model $\morep$ to be $\loss_m / \loss_{\hat{\mo}}$. 
Note that the concepts of efficiency and asymptotic efficiency in model selection are reminiscent of their counterparts in parameter estimation.  
We plot the efficiency of each candidate model in Figure~\ref{fig:forSlides_AR3}(c).
Similarly to Figure~\ref{fig:forSlides_AR3}(b), the curves here show that the true model is the most efficient model. 
We note that the minus-$\sigma^2$ adjustment of out-sample prediction loss in the above definition makes the property highly nontrivial to achieve (see for example~\cite{shibata1980asymptotically,shao1997asymptotic,yang2006comparing}). Consider for example the comparison between AR(2) and AR(3) models, with the AR(2) being the true data-generating model. It can be proved that without subtracting $\sigma^2$, the ratio (of the mean square prediction errors) for each of the two candidate models approaches $1$; by subtracting $\sigma^2$, the ratio for AR(2) still approaches $1$ while the ratio for AR(3) approaches $2/3$.

\begin{figure*}[tb]
\centering
  \includegraphics[width=0.9\linewidth]{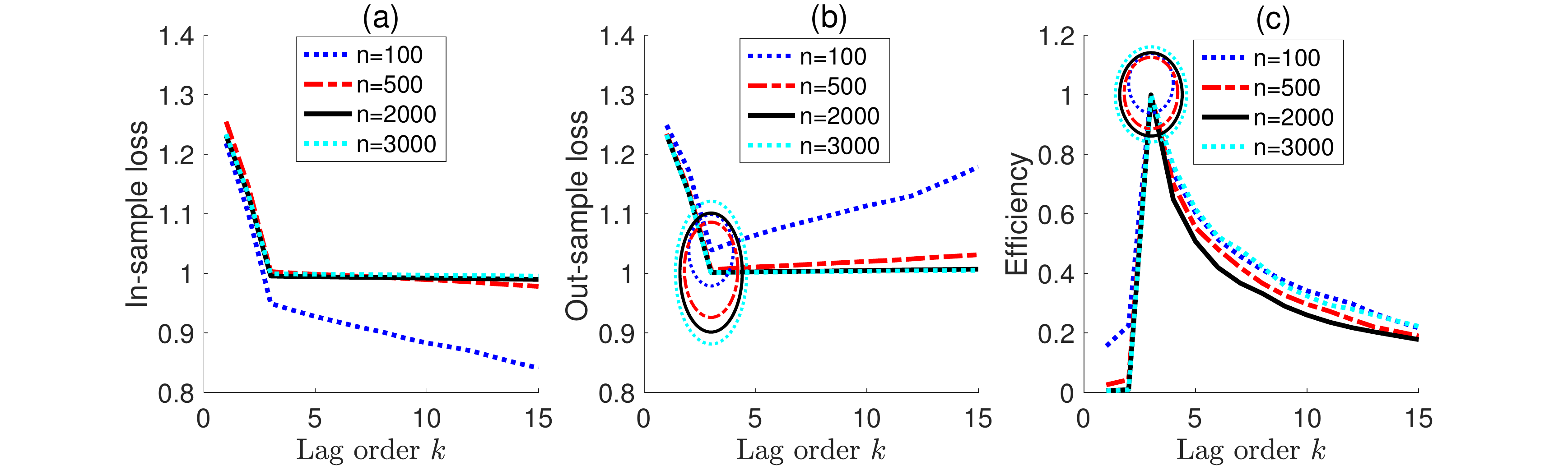}
  \vspace{-0.1in}
  \caption{Parametric framework: the best predictive performance is achieved at the true order 3 }
  \label{fig:forSlides_AR3}
  \vspace{-0.1in}
\end{figure*}

2) Nonparametric framework:
The data are generated by the moving average model 
$z_t = \v_t - 0.8 \v_{t-1}$, with $\v_t$ being independent standard Gaussian. 

Similarly to 1), we plot the results in Figure~\ref{fig_forSlides_MA1}.  
Different from case 1), the predictive performance is  optimal at increasing model dimensions as $n$ increases. In such a nonparametric framework, the best model is  sensitive to the sample size, so that pursuing an inference of a fixed good model becomes {\coco unrealistic}. 
The model selection task aims to select a model that is asymptotically efficient (see Figure~\ref{fig_forSlides_MA1}(c)). 
{\coco Note that Figures~\ref{fig_forSlides_MA1}(b)(c) are drawn} based on the information of the underlying true model which is unavailable in practice, hence we need a model selection method to achieve the asymptotically efficiency.

\begin{figure*}[tb]
\centering
  \includegraphics[width=0.9\linewidth]{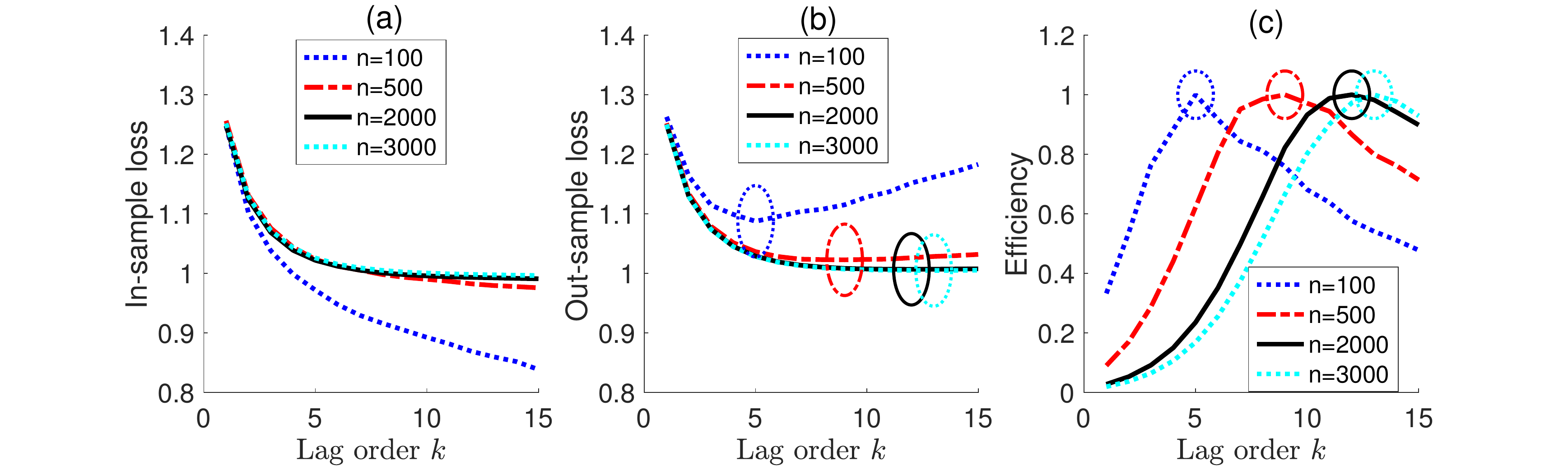}
  \vspace{-0.1in}
  \caption{Nonparametric framework: the best predictive performance  is achieved at an order that depends on the sample size
  }
  \label{fig_forSlides_MA1}
  \vspace{-0.1in}
\end{figure*}

3) Practically nonparametric framework:
The data are generated in the way as in 1), except that $k_0=10$. 

We plot the results in Figure~\ref{fig:forSlides_AR10}.  
For $n=2000,3000$, the sample sizes are large enough to support the evidence of a true model with a relatively small model dimension. Similarly to experiment 1), this is a parametric framework in which the optimal predictive performance is achieved at the true model. 
For $n=100,500$, where the sample sizes are not large enough compared to the true model dimension, however,  fitting too  many parameters actually causes an increased 
variance which diminishes the predictive power. In such a scenario, even though the true model is included as a candidate, the best model is not the true model and it is unstable for small or moderate sample sizes as if in a nonparametric setting.  
In other words, the parametric framework can turn into a practically nonparametric framework in the small data regime. 
{\coco It can also work the other way around, i.e., for a true nonparametric framework, for a large range of sample sizes (e.g., 100 to 2000), a relatively small parametric model among the candidates maintains to be the best model \cite{liu2011parametric}.} 
 
\begin{figure*}[tb]
\centering
  \includegraphics[width=0.9\linewidth]{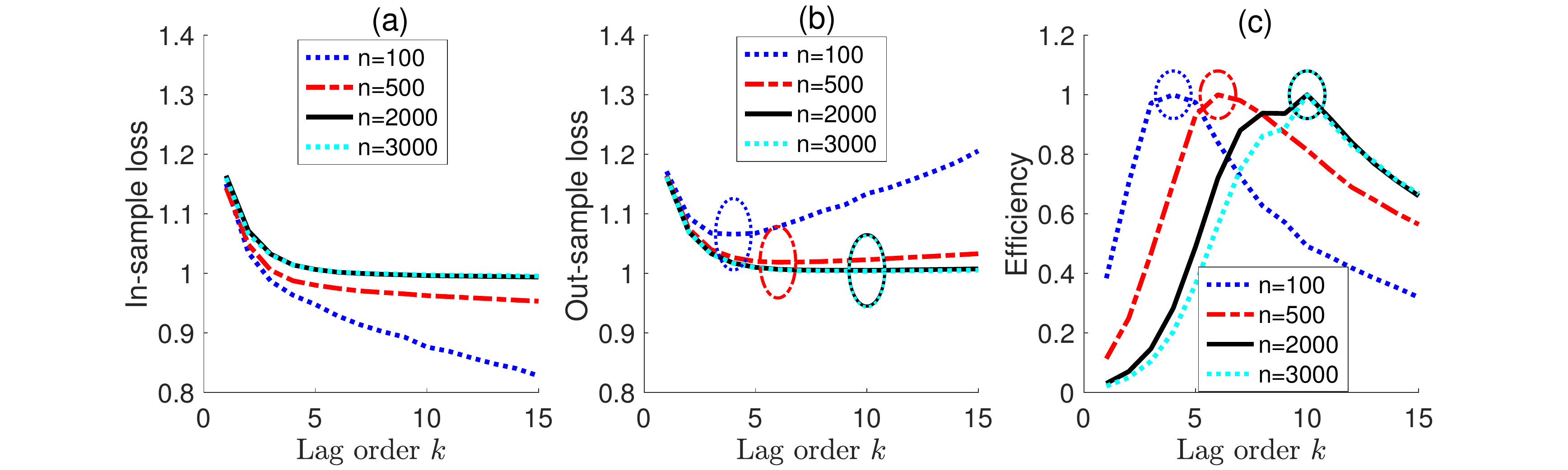}
  \vspace{-0.1in}
  \caption{Practically nonparametric framework: the best predictive performance is achieved at an order that depends on the sample size in the small data regime} 
  \label{fig:forSlides_AR10}
  \vspace{-0.2in}
\end{figure*}
 

}


\section{Principles and Approaches from Various Philosophies or Motivations} \label{sec:criteria}


A wide variety of model selection methods have been proposed in the past few decades, motivated by different viewpoints and justified under various circumstances.  
Many of them originally aimed to select either the order in an AR model or subset of variables in a regression model. 
We review some of the representative approaches in these contexts in this section. 

\subsection{Information criteria based on likelihood functions} \label{subsec:IC}

Information criteria generally refer to model selection methods that are based on likelihood functions and applicable to  parametric model based problems. {\coco Here we introduce some information criteria whose asymptotic performances are well understood.}

\textbf{Akaike information criterion} (AIC) is a model selection principle proposed by Akaike\cite{akaike1974new}. 
A detailed derivation of it from an information theoretic perspective can be found in \cite{stoica2004model}. Briefly speaking, the idea is to approximate the out-sample prediction loss by the sum of the in-sample prediction loss and a correction term. We refer to \cite{stoica2004model} for a detailed derivation of this correction term. In the typical setting where the loss is logarithmic, the AIC procedure is to select the model $\morep$ that minimizes 
\begin{align}
	\AIC_m = -2 \hat{\ell}_{n,m} + 2 \d, \label{eq:AIC}
\end{align}
{\coco where $\hat{\ell}_{n,m}$ is the maximized log-likelihood of model $\morep$ given $n$ observations as defined in (\ref{eq:MLE}), and $d_m$ is the dimension of model $\morep$.} 
It is clear that more complex models (with larger $d_m$) will suffer from larger penalties. 

In the task of autoregressive order selection,  it is also common to use 
\begin{align}
\AIC_k = n \log \hat{e}_k + 2 k \label{eq:AIC2}
\end{align}
for the model of order $k$, 
where $\hat{e}_k$ is the average in-sample prediction error based on the quadratic loss. In fact, (\ref{eq:AIC2}) can be derived from (\ref{eq:AIC}) by assuming that AR noises are Gaussian and by regarding ARs of different orders as $\Mo$.
A predecessor of AIC is the final prediction error criterion (FPE)~\cite{akaike1969fitting} (also by Akaike).  
An extension of AIC is the Takeuchi's information criterion~\cite{TIC} derived in a way that allows model mis-specification, but it is rarely used in practice due to its computational complexity. 
In the context of generalized estimating equations for correlated response data, a variant of AIC based on quasi-likelihood is derived in~\cite{pan2001akaike}.

\textbf{Finite-sample corrected AIC} (AICc)~\cite{hurvich1989regression} was proposed as a corrected version of the AIC for small-sample study. 
It selects the model that minimizes
$$
\AICc_m = \AIC_m + \frac{2 \{\d+1\} \{\d+2\}}{n-\d-2} .
$$
Unless the sample size $n$ is small compared with model dimension $\d$, there is little difference between $\AICc$ and $\AIC$.
Another modified AIC that replaces the constant 2 with a different positive number has also been studied in \cite{broersen2000finite}.

\textbf{Bayesian information criterion} (BIC)~\cite{schwarz1978estimating} is another popular model selection principle. It selects the model $m$ that minimizes 
\begin{align}
	\BIC_m = -2 \hat{\ell}_{n,m} +  \d  \log n. \label{eq:bic}
\end{align}
The only difference with AIC is that the constant $2$ in the penalty term is replaced with the logarithm of the sample size. The original derivation of BIC by Schwarz turned out to have a nice Bayesian interpretation, as its current name suggests. 

To see the interpretation, we assume that $z_1,\ldots,z_n$ are the realizations of i.i.d. random variables, and $\pi(\cdot)$ is any prior distribution on $\theta$ which has dimension $d$. We let $\ell_n(\theta) = \sum_{i=1}^n \log p_{\theta}(z_i)$ be the log-likelihood function, and $\hat{\th}_n$ the MLE of $\th$. A variant of the Bernstein-von Mises theorem~\cite[Chapter 10.2]{van1998asymptotic} gives (under regularity conditions) 
\begin{align}
	&\int_{\mathbb{R}^d}  \pi(\hat{\theta}_n+n^{-\frac{1}{2}} r) \exp \biggl( \ell_n(\hat{\theta}_n+n^{-\frac{1}{2}} r) - \ell_n( \hat{\theta}_n ) \biggr) dr  \nonumber \\
	&\limp \pi(\th_*) (2\pi)^{d/2} \{ \det \E(-\nabla_{\theta}^2 \log p_{\th_*}(z)) \}^{-1/2} \label{eq:von}
\end{align}
as $n \rightarrow \infty$, for some constant $\th_*$. Note that the right hand side of (\ref{eq:von}) is a constant that does not depend on $n$. 
Direct calculations show that the left hand side of (\ref{eq:von}) equals 
\begin{align}
p(z_1,\ldots, z_n)\exp\biggl( -\ell_n(\hat{\theta}_n) + \frac{d}{2} \log n \biggr).\label{eq:concerntration}
\end{align}
Therefore, selecting a model with the largest marginal likelihood $p(z_1,\ldots, z_n)$ (as advocated by Bayesian model comparison) is asymptotically equivalent to selecting a model with the smallest BIC in (\ref{eq:bic}). It is interesting to see that the marginal likelihood of a model does not depend on the imposed prior at all {\coco in the large sample limit}. Intuitively speaking, this is because in the integration of $p(z_1,\ldots, z_n) = \int_{\th} \pi(\th) p_{\th}(z_1,\ldots, z_n) d \th$, the mass is concentrated around $\hat{\th}_n$ with radius $O(n^{-1/2})$ and dimension $d$, so its value is proportional to the maximized likelihood value multiplied by the volume approximately at the order of $n^{-d/2}$, which is in line with (\ref{eq:concerntration}).

\textbf{Hannan and Quinn criterion} (HQ)~\cite{hannan1979determination} was proposed as an information criterion that achieves  strong consistency in autoregressive order selection.
In other words, if the data are truly generated by an autoregressive model of fixed order $k_0$, then the selected order $k$ converges almost surely to $k_0$ as the sample size goes to infinity. 
{\coco We note that strong consistency implies (the usual) consistency. }
In general, this method selects a model by minimizing  
$ \HQ_m = -2 \hat{\ell}_{n,m} + 2 c \ d_m \log\log n$ (for any constant $c>1$).  It can be proved under some conditions that any penalty no larger than $2 d_m \log\log n$ is not strongly consistent~\cite{hannan1979determination}; therefore, HQ employs the smallest possible penalty to guarantee strong consistency.

\textbf{Bridge criterion} (BC)~\cite{DingBridge,DingBridge2} is a recently proposed information criterion that aims to bridge the advantages of both AIC and BIC in the asymptotic regime. It selects the model $\morep$ that minimizes $ \BC_m = -2 \hat{\ell}_{n,m} + c_n (1+2^{-1}+\cdots + \d^{-1})$ (with the suggested $c_n=n^{2/3}$) over all the candidate models whose dimensions are no larger than $d_{m_{\AIC}}$, the dimension of the model selected by $\AIC$. Note that the penalty is approximately $c_n \log d_m$, but it is written as a harmonic number to highlight some of its nice interpretations. 
Its original derivation was motivated by a recent finding that the information loss of underfitting a model of dimension $d$ using dimension $d-1$ is asymptotically $\chi_1^2/d$ for large $d$, assuming that nature generates the model from a non-informative uniform distribution over its model space (in particular the coefficient space of all stationary autoregressions)~\cite[Appendix A]{DingBridge}. BC was proved to perform similarly to AIC in a nonparametric framework, and similarly to BIC in a parametric framework. {\coco We further discuss BC in Section~\ref{sec:AICBIC}.}

\subsection{Methods from other perspectives}

In addition to information criteria, some other model selection approaches have been motivated from either Bayesian, information-theoretic, or decision-theoretic perspectives. 

\textbf{Bayesian posterior probability} is commonly used in Bayesian data analysis. Suppose that  each model $\mo \in \moind$ is assigned a prior probability $p(\morep)>0$ (such that $\sum_{m\in \moind}p(\morep) = 1$), interpreted as the probability that model $\morep$ contains the true data-generating distribution $p_*$. Such a prior may be obtained from scientific reasoning or {\coco knowledge from} historical data. For each $m \in \moind$ we also introduce a prior with density $\th_m \mapsto p_m(\th_m)$ ($\th_m \in \H$), and a likelihood of data $p_m(z \mid \th_m)$ where $z=[z_1,\ldots,z_n]$. A joint distribution on $(z,\th_m,\morep)$ is therefore well defined, based on which various quantities of interest can be calculated. We first define the marginal likelihood of model $\morep$ by
\begin{align} 
	p(z \mid \morep) = \int_{\H} p_m(z \mid \th_m) p_m(\th_m) d \th_m	\label{eq:marginal}.
\end{align}
Based on (\ref{eq:marginal}), we obtain the following posterior probabilities on models by Bayes formula
\begin{align}
p(\morep \mid z) = \frac{p(z \mid \morep ) p(\morep)}{\sum_{m' \in \moind } p(z \mid \mathcal{M}_{m'}) p(\mathcal{M}_{m'})}.\label{eq:BF} 
\end{align}
The maximum a posteriori approach~\cite{djuric1998asymptotic} would select the  model with the largest posterior probability. 

\textbf{Bayes factors} are also popularly adopted  for  Bayesian model comparison, defined for a pair of models $(\mathcal{M}_{m}, \mathcal{M}_{m'})$ by 
$$
B_{m,m'} = \frac{p(\mathcal{M}_{m} \mid z)}{p(\mathcal{M}_{m'} \mid z)} / \frac{p(\mathcal{M}_{m})}{p(\mathcal{M}_{m'})} 
= \frac{p(z \mid \mathcal{M}_{m})}{p(z \mid \mathcal{M}_{m'})} .
$$
The model $\mathcal{M}_{m}$ is favored over $\mathcal{M}_{m'}$ if $B_{m,m'} > 1$. 
Bayes factors remove the impact of prior probabilities on the models from the selection process, to focus on the ratio of marginal likelihoods. 
Compared with the Bayesian posterior probability, Bayes factors are appealing when it is difficult to formulate prior probabilities on models. 

\textbf{Bayesian marginal likelihood} defined in (\ref{eq:marginal}), also referred to as the evidence or model evidence, is a quantity naturally motivated by Bayes factors. In the presence of multiple models, the one with the largest Bayesian marginal likelihood is favored over all other models in terms of the Bayes factor. 
Moreover, it can be seen that the model with the highest marginal likelihood is the model with the highest posterior probability given that the Bayesian prior probabilities on models are all equal.
Interestingly, this Bayesian principle using marginal likelihood is asymptotically equivalent to the BIC (as we have introduced in Subsection~\ref{subsec:IC}).  
In practice, the above Bayesian model selection methods can be computationally challenging. Calculation of the quantities in (\ref{eq:marginal}) and (\ref{eq:BF}) are usually implemented using Monte Carlo methods, especially sequential Monte Carlo (for online data) and Markov chain Monte Carlo (for batch data)~(see, e.g.,~\cite{andrieu2001model}). 
It is worth noting that improper or vague priors on the parameters of any candidate model can have non-negligible impact on the interpretability of marginal likelihood and Bayes factors in the non-asymptotic regime, and that has motivated some recent research on Bayesian model selection~\cite{shao2017bayesian}.

\textbf{Minimum message length} (MML) principle \cite{wallace1968information} was proposed from an  information-theoretic perspective. It favors the model that generates the shortest overall message, which consists of a statement of the model and a statement of the data concisely encoded with that model. Specifically, this criterion aims to select the model that minimizes 
$$
-\log p( \theta) - \log p(x \mid  \theta) + \frac{1}{2}\log | I( \theta) | + \frac{d}{2} (1 + \log \kappa_{d} ) .
$$ 
where $p(\theta)$ is a prior, $p(x \mid \theta)$ is the likelihood function, $I(\theta)=\int \{\partial \log p(x \mid \theta) / \partial \theta\}^2 p(x \mid \theta) dx$ is the  Fisher information, $d$ is the dimension of $\th$, and $\kappa_{d}$ is the so-called optimal quantizing lattice constant that 
is usually approximated by $\kappa_{1}=1/12$. 
A detailed derivation and application of MML can be found in \cite{figueiredo2002unsupervised}. 

\textbf{Minimum description length} (MDL) principle~\cite{rissanen1978modeling,Rissanen1982,barron1998minimum,hansen2001model}  
	describes the best model as the one that leads to the best compression of a given set of data. 
It was also motivated by an information-theoretic perspective (which is similar to MML). 
Different with MML which is in a fully Bayesian setting, MDL avoids assumptions on prior distribution. 
Its predictive extension, referred to as the predictive minimum description length  criterion (PMDL), is  proposed in \cite{rissanen1986stochastic}.
One formulation of the principle is to select the model by minimizing the  stochastic complexity 
$-\log p_{ \theta_1}(z_1) - \sum_{t=2}^n \log p_{ \theta_t}(z_t \mid z_1,\ldots, z_{t-1}),$ in which $\theta_t$'s are restricted to the same parameter space (with the same dimension).  
Here, each $ \theta_t$ ($t>1$) is the MLE calculated using $z_1,\ldots,z_{t-1}$ and $p_{ \theta_1}(\cdot)$ can be an arbitrarily chosen prior distribution. 
The above PMDL criterion is also closely related to the   prequential (or predictive sequential) rule \cite{dawid1984present} from a decision-theoretic perspective.

\textbf{Deviance information criterion} (DIC)~\cite{spiegelhalter2002bayesian} was derived as a measure of Bayesian model complexity. Instead of being derived from a frequentist perspective, DIC can be thought of as a Bayesian counterpart of AIC. To define DIC, a relevant concept is the deviance under model $\mo$: $D_{\mo}(\theta)=-2\log p_{\mo}(y|\theta )+C$ where $C$ 
 does not depend on the model being compared. 
Also we define the ``effective number of parameters'' of the model to be  
$p_{D}=E_{\th \mid z} D_{\mo}(\theta) - D_{\mo}(E_{\th \mid z} (\th))$, 
where $E_{\th \mid z}(\cdot)$ is the expectation taken over $\theta$ conditional on all the observed data $z$ under model $\morep$. 
Then the DIC selects the model $\morep$ that minimizes
\begin{align}
	\DIC_{\mo} = D_{\mo}(E_{\th \mid z} (\th) ) + 2 p_D. \label{eq:dic}
\end{align}	 
The conceptual connection between DIC and AIC can be readily observed from (\ref{eq:dic}).
The MLE and model dimension in AIC are replaced with the posterior mean and effective number of parameters respectively in DIC.
Compared with AIC, DIC enjoys some computational advantage for comparing complex models whose likelihood functions may not even be in analytic forms. In Bayesian settings, Markov chain Monte Carlo tools can be utilized to simulate  posterior distributions of each candidate model, which can be further used to efficiently compute DIC in (\ref{eq:dic}).

\subsection{Methods that do not require parametric assumptions}

\textbf{Cross-validation} (CV) 
\cite{allen1974relationship,geisser1975predictive} is a class of model selection methods widely used in machine learning practice. 
CV does not require the candidate models to be parametric, and it works as long as the data are permutable and one can assess the predictive performance based on some measure.
A specific type of CV is the delete-$1$ CV method \cite{stone1977asymptotic} (or leave-one-out, LOO). 
The idea is explained as follows. For brevity, let us consider a parametric model class as before. Recall that we wish to select a model $\morep$ with as small out-sample loss $E_*(s(p_{\tilde{\th}_m}, Z))$ as possible. Its computation involves an unknown true data-generating process, but we may approximate it by 
$n^{-1} \sum_{i=1}^n s(p_{\hat{\th}_{m,-i}}, z_i)$
where $\hat{\th}_{m,-i}$ is the MLE under model $\morep$ using all the observations except $z_i$. 
In other words, given $n$ observations, we leave each one observation out in turn and attempt to predict that data point by using the $n-1$ remaining observations, and record the average prediction loss over $n$ rounds.
Interestingly, the LOO was shown to be asymptotically equivalent to either AIC or TIC under some regularity conditions~\cite{stone1977asymptotic}. 

In general, CV works in the following way. It first randomly splits the original data into a training set of $n_t$ data $1 \leq n_t \leq n-1$ and a validation set of $n_v = n-n_t$ data; each candidate model is then trained from the $n_t$ data and validated on the remaining data (i.e. to record the average validation loss); the above procedure is independently replicated a few times (each with a different validation set) in order to reduce the {\coco variability caused by splitting}; finally, the model with the smallest average validation loss is selected, and it is re-trained using the complete data for future prediction.  

A special type of CV is the so-called $k$-fold CV (with $k$ being a positive integer). It randomly partitions data into $k$ subsets of (approximately) equal size; each model is trained on $k-1$ folds and validated on the remaining $1$ fold; the procedure is repeated $k$ times, and the model with the smallest  average validation loss is selected. 
The $k$-fold CV is perhaps more commonly used than LOO, partly due to the large computational complexity involved in LOO. 
The holdout method, as often used in data competitions (e.g., Kaggle competition) is also a special case of CV: It does data splitting only once, one part as the training set and the remaining part as the validation set.
{\coco We note that there exist fast methods such as generalized cross-validation (GCV) as surrogates to LOO in order to reduce the computational cost.}
Some additional discussion on CV will be elaborated on in Section~\ref{sec:clarify}.

\subsection{Methods proposed for specific types of applications}

There have been some other criteria proposed for specific types of applications, mostly for time series or linear regression models. 

\textbf{Predictive least squares} (PLS) principle proposed by Rissanen~\cite{rissanen1986predictive} is a model selection criterion  based on his PMDL principle. PLS aims to select the stochastic regression model by minimizing the accumulated squares of prediction errors (in a time series setting), defined as 
$$
\PLS_{\mo}  = \sum_{t=t_0+1}^n (y_t -  x_{\mo,t}^\T  \beta_{m,t-1} )^2
$$
where $y_t$ is each response variable, $ x_{\mo,t}$ is the vector of covariates corresponding to model $\mo$, and $ \beta_{m,t-1}$ is the least squares estimate of model $\morep$ based on data before time $t$. The time index $t_0$ is the first index such that $\beta_{t}$ is uniquely defined.  Conceptually, PLS is not like AIC and BIC that select the model that minimizes a loss plus a penalty; it seems more like the counterpart of LOO in sequential contexts. Interestingly, it has been proved that PLS and BIC are asymptotically close, both strongly consistent in selecting the data-generating model (in a parametric framework)~\cite{wei1992predictive}. Extensions of PLS where the first index $t_0$ is a chosen sequence indexed by $n$ have also been studied. For example, it has been shown under some conditions that PLS with $t_0/n \rightarrow 1$ shares the same asymptotic property of AIC (see, e.g.,~\cite{ing2007accumulated} and the references therein).

\textbf{Generalized information criterion} (GIC$_{\lambda_n}$)
\cite{nishii1984asymptotic,shao1997asymptotic}
represents a wide class of criteria whose penalties are linear in model dimension. It aims to select the regression model $\morep$ that minimizes 
$$
\textsc{gic}_{\lambda_n,m} = \hat{e}_{\mo} + \frac{\lambda_n \hat{\sigma}_n^2  \d}{n}.
$$
Here, $\hat{\sigma}_n^2$ is an estimator of  $\sigma^2$, the variance of the noise, and $\hat{e}_m = n^{-1} \norm{y - \hat{y}_m}_2^2$ is the mean square error between the observations and least squares estimates under regression model $\morep$. $\lambda_n$ is a deterministic sequence of $n$ that controls the trade-off between the model fitting and  model complexity.
If  we replace $\hat{\sigma}_n^2$ with $(n-\d)^{-1} n \hat{e}_{\mo}$, it can be shown under mild conditions that minimizing $\GIC$ is equivalent to minimizing~\cite[pp. 232]{shao1997asymptotic}  
\begin{align}
	\log  \hat{e}_{\mo} + \frac{\lambda_n \d}{n}.
\end{align}	
In this case, $\lambda_n=2$ corresponds to AIC and $\lambda_n=\log n$ corresponds to BIC. 
\textbf{Mallows' $C_p$ method}~\cite{mallows1973some} is a special case of GIC with $\hat{\sigma}_n^2 \de (n-d_{\bar{m}})^{-1} n \hat{e}_{\bar{m}}$ and $\lambda_n = 2$, where $\bar{m}$ indexes the largest model which includes all the covariates.

\subsection{Theoretical properties of the model selection criteria} \label{subsec:theory}

Theoretical examinations of model selection criteria have centered on several properties: consistency in selection, asymptotic efficiency and minimax-rate optimality. Selection consistency targets the goal of identifying the best model or method on its own for scientific understanding, statistical inference, insight or interpretation. Asymptotic efficiency and minimax-rate optimality (defined in Definition~\ref{def:minimax} below) are in tune with the goal of prediction. 
Before we introduce the theoretical properties, it is worth mentioning that many model selection methods can also be categorized into two classes according to their large-sample performances,  respectively represented by AIC and BIC. 
In fact, it has been known that AICc, FPE and GCV are asymptotically close to AIC;  
on the other hand, Bayes factors, HQ, and the original PLS are asymptotically close to BIC. 
For some other methods such as CV and GIC, their asymptotic behavior usually depends on the tuning parameters. 
GIC$_{\lambda_n}$ is asymptotically equivalent to AIC when $\lambda_n=2$ and to BIC when $\lambda_n=\log n$. In general, any sequence of $\lambda_n$ satisfying $\lambda_n \rightarrow \infty$ would exhibit the consistency property shared by BIC.
As a corollary, the $C_p$ method (as a special case of GIC$_2$) is asymptotically equivalent to AIC.  
For CV with $n_t$ training data and $n_v$ validation data, it is asymptotically similar to AIC when $n_v/n_t \rightarrow 0$ (including the LOO as a special case), and to BIC when $n_v/n_t \rightarrow \infty$~\cite[Eq. (4.5)]{shao1997asymptotic}.

In general, AIC and BIC have served as the golden rules for model selection in statistical theory since their existence.   
Though cross-validations or Bayesian procedures have also been widely used, their asymptotic justifications are still rooted in frequentist approaches in the form of AIC, BIC, etc. 
Therefore, understanding the asymptotic behavior of AIC and BIC is crucial in both theory and practice.  
We therefore focus on the properties of AIC and BIC in the rest of this section and Section~\ref{sec:AICBIC}. It is remarkable that the asymptotic watershed of AIC and BIC (and their closely related methods) simply lies in whether the penalty is a fixed well-chosen constant or goes to infinity as a function of $n$.

First of all, AIC is proved to be \textit{minimax-rate optimal} for a range of variable selection tasks, including the usual subset selection and order selection problems in linear regression, and nonparametric regression based on series expansion with basis such as polynomials, splines, or wavelets (see, e.g.,~\cite{barron1999risk} and the references therein). For example, consider the minimax risk of estimating the regression function $f\in \mathcal{F}$ under the squared error
\begin{align}
	\inf_{\hat{f}} \sup_{f \in \mathcal{F}} n^{-1} \sum_{i=1}^n \E (\hat{f}(x_i)-f(x_i))^2,\label{eq:uniform}
\end{align}
where $\hat{f}$ is over all estimators based on the observations, and $f(x_i)$ is the expectation of the $i$th response variable (or the $i$th value of the regression function) conditional on the $i$th vector of variables $x_i$. 
Each $x_i$ can refer to a vector of explanatory variables, or polynomial basis terms, etc. 
{\coco For a model selection method $\nu$}, its worst-case risk is $\sup_{f \in \mathcal{F}} R(f,\nu, n)=n^{-1} \sum_{i=1}^n \E \{\hat{f}_{\nu}(x_i)-f(x_i)\}^2$ with $\hat{f}_{\nu}$ being the least squares estimate of $f$ under the variables selected by $\nu$. 
\begin{definition} \label{def:minimax}
A method $\nu$ 
  is said to be minimax-rate optimal over $\mathcal{F}$ if 
$\sup_{f \in \mathcal{F}} R(f,\nu, n)$ converges at the same rate as the minimax risk in (\ref{eq:uniform}).
\end{definition}

Another good property of AIC is that it is \textit{asymptotically efficient} (as defined in (\ref{eq:generalEfficiency})) in a nonparametric framework (see, e.g.,~\cite{shibata1980asymptotically,shibata1981optimal}). In other words, the predictive performance of its selected model is asymptotically equivalent to the best offered by the candidate models (even though it is  sensitive to the sample size).

BIC, on the other hand, is known to be consistent in selecting the smallest true data-generating model in a parametric framework (see, e.g.,~\cite{hannan1979determination,shao1997asymptotic}). For example, suppose that the data are truly generated by an AR(2), and the candidate models are AR(2), AR(3), and a moving average model which is essentially AR($\infty$). Then  AR(2) is selected with probability going to one as the sample size tends to infinity. MA(1) is not selected because it is a wrong model, and AR(3) is not selected because it overfits (even though it nests AR(2) as its special case).  
Moreover, it can be proved that the consistency of BIC also implies that it is \textit{asymptotically efficient} in a parametric framework~\cite{shao1997asymptotic,DingBridge}. 
We will elaborate more on the theoretical properties of AIC and BIC in Section~\ref{sec:AICBIC}.


\section{War and Peace---Conflicts Between AIC and BIC, and Their Integration}
\label{sec:AICBIC}

In this section, we review some research advances in the understanding of AIC, BIC, and related criteria. The choice of AIC and BIC to focus on here is because they represent two cornerstones of model selection principles and theories.
We are only concerned with the 
settings where the sample size is larger than the model dimension. 
Details of the following discussions can be found in original papers such as \cite{shibata1976selection,shibata1980asymptotically,shao1997asymptotic,shibata1981optimal,yang2005can,DingBridge} and the references therein.

Recall that AIC is asymptotically efficient for the nonparametric framework and is also minimax optimal~\cite{barron1999risk}. In contrast, BIC is consistent and asymptotically efficient for the parametric framework.
Despite the good properties of AIC and BIC, they have their own drawbacks. 
AIC is known to be inconsistent in a parametric framework where there are at least two correct candidate models. As a result, AIC is not asymptotically efficient in such a framework.  For example, if data are truly generated by an AR(2), and the candidate models are AR(2), AR(3), etc., then AR(2) cannot be selected with probability going to one by AIC as the sample size increases. The asymptotic probability of it being selected can actually be analytically computed~\cite{shibata1976selection}.  
BIC, on the other hand, does not enjoy the properties of minimax-rate optimality and asymptotic efficiency in a nonparametric framework~\cite{foster1994risk,shao1997asymptotic}. 

Why do AIC and BIC work in those ways? In fact, theoretical arguments in those aspects are highly nontrivial and have motivated a vast literature since the formulations of AIC and BIC.  Here we provide some heuristic explanations. For AIC, its formulation in (\ref{eq:AIC}) was originally motivated by searching the candidate model $p$ that is the closest in Kullback-Leibler (KL) divergence (denoted by $D_{KL}$) from $p$ to the data-generating model $p_*$. Since $\min_{p} D_{KL}(p_*,p)$ is equivalent to $\min_{p} \E (-\log p)$ for a fixed $p_*$, AIC is expected to perform well in minimizing the prediction loss. But AIC is not consistent for a model class containing a true model and at least one oversized model, because fitting the oversized model would only reduce the first term $-2 \hat{\ell}_{n,m}$ in (\ref{eq:AIC}) by a random quantity that is approximately chi-square distributed (by, e.g., Wilks' theorem~\cite{wilks1938large}), while the increased penalty on the second item $2\d$ is at a constant level which is not large enough to suppress the overfitting gain in fitness with high probability. 
On the other hand, selection consistency of BIC in a parametric framework is not surprising due to its nice Bayesian interpretation (see Section~\ref{sec:criteria}).   However, its penalty $\d \log n$ in (\ref{eq:bic}) is much larger than the $2 \d $ in AIC, so it cannot enjoy the predictive optimality in a nonparametric framework (if AIC already does so).

\begin{figure}
\centering
  \includegraphics[angle=0,width=1\linewidth]{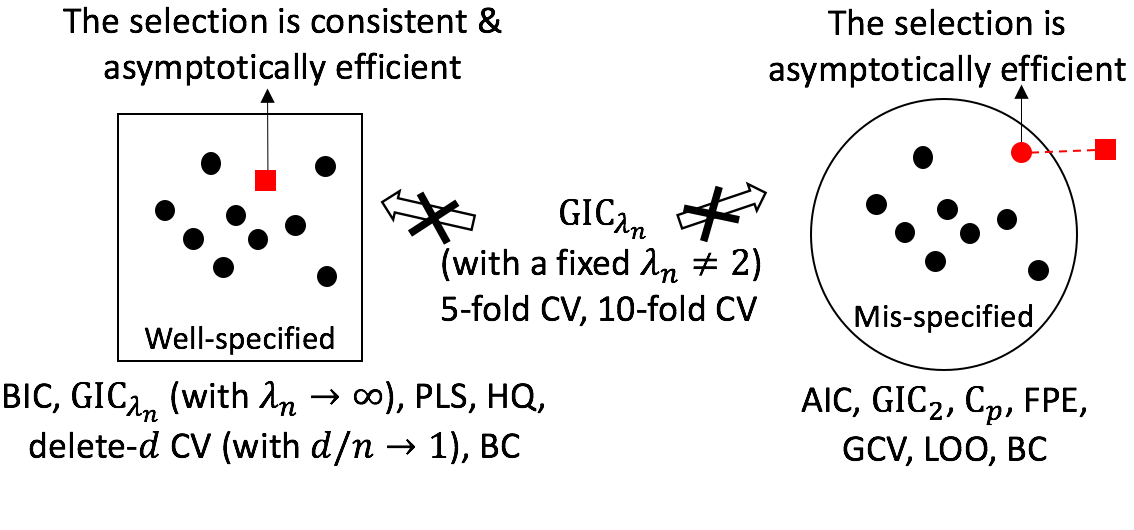}
  \vspace{-0.2in}
  \caption{A graph illustrating a parametric setting where the model class (by large square) includes the true data generating model (by small red square), a nonparametric setting where the model class (by large circle) excludes the true data generating model, along with an asymptotically efficient model (by red circle) in the second case. It also lists some popular methods suitable for either situation, and a class of GIC and CV that are asymptotically sub-optimal for regression models. }
  \label{fig:efficiency}
  \vspace{-0.2in}
\end{figure}

To briefly summarize, for asymptotic efficiency, AIC (resp. BIC) is only suitable in nonparametric (resp. parametric) settings.
Figure~\ref{fig:efficiency} illustrates the two situations.
There has been a debate between AIC and BIC in model selection practice, centering on whether the data generating process is in a parametric framework or not. 
The same debate was sometimes raised under other terminology. In a parametric (resp. nonparametric) framework, the true data-generating model is often said to be well-specified (resp. mis-specified), or finitely (resp. infinitely) dimensional.\footnote{To see a reason for such terminology, consider for instance the regression analysis using polynomial basis function as covariates. If the true regression function is indeed a polynomial, then it can be parameterized with a finite number of parameters; if it is an exponential function, then it cannot be parameterized with any finitely dimensional parameter. }
Without prior knowledge on how the observations were generated, determining which method to use becomes very challenging. 
It naturally motivates the following fundamental question:  

``\textit{Is it possible to have a method that combines the strengths of AIC and BIC}?'' 
 
The combining of strengths can be defined in two ways. 
Firstly, can the properties of \textbf{minimax-rate optimality and consistency} be shared?
Unfortunately, it has been theoretically shown under rather general settings that there exists no model selection method that achieves both optimality simultanously~\cite{yang2005can}.
That is, for any model selection procedure to be consistent, it must behave sup-optimally in terms of minimax rate of convergence in the prediction loss.  
Secondly, can the properties of \textbf{asymptotic efficiency and consistency} be shared? 
In contrast to minimax-rate optimality  which allows the true data-generating model to vary, asymptotic efficiency is in a pointwise sense, meaning that the data are already generated by some fixed (unknown) data-generating model. 
Therefore, the asymptotic efficiency is a requirement from a more optimistic view and thus weaker in some sense than the  minimaxity.  
Recall that consistency in a parametric framework is typically equivalent to asymptotic efficiency~\cite{shao1997asymptotic,DingBridge}. 
Clearly, if an ideal method  can combine  asymptotic efficiency and consistency, it  achieves asymptotic efficiency in both parametric and nonparametric frameworks. 
That motivated an active line of recent advances in reconciling the two classes of model selection methods~\cite{ing2007accumulated,erven2012catching,DingBridge}. 

In particular, a new model selection method called Bridge criterion (BC) was recently proposed (see Section~\ref{sec:criteria}) to simultaneously achieve consistency in a parametric framework and asymptotic efficiency in both (parametric and nonparametric) frameworks.
{\coco The key idea of BC  
is to impose a BIC-like heavy penalty for a range of small models, but to alleviate the penalty for larger models if more evidence is supporting an infinitely dimensional true model. In that way, the selection procedure is automatically adaptive to the appropriate setting (either parametric or nonparametric).
A detailed statistical interpretation of how BC works in both theory and practice, and how it relates to AIC and BIC are elaborated in \cite{DingBridge}.}

Moreover, in many applications, data analysts would like to quantify to what extent the framework under consideration can be practically treated as parametric, or in other words, how likely the postulated model class is well-specified. This motivated the concept of ``parametricness index'' (PI)~\cite{liu2011parametric,DingBridge} which assigns a confidence score to model selection. One definition of PI, which we shall use in the following experiment, is the following quantity on $[0,1]$:
$$
\PI = |d_{m_{\BC}} - d_{m_{\AIC}}|/(|d_{m_{\BC}} - d_{m_{\AIC}}| + |d_{m_{\BC}} - d_{m_{\BIC}}|)
$$
if the denominator is not zero, and $\PI = 1$ otherwise.  
Here, $d_{m_{\nu}}$ is the dimension of the model selected by the method $\nu$. 
Under some conditions, it can be proved that $\PI \limp 1$ in a parametric framework and $\PI \limp 0$ otherwise.

\begin{table}[tb]
\centering
\caption{Autoregressive order selection: the average efficiency, dimension, and PI (along with standard errors)}
\label{table:AR}
\begin{tabular}{ccccc}
\hline
                         &                          & AIC         & BC          & BIC                 \\ \hline
\multirow{3}{*}{Case 1)} & Efficiency  & 0.78 (0.04) & \textbf{0.93} (0.02) & \textbf{0.99} (0.01)  \\ \cline{2-5} 
                         & Dimension        & 3.95 (0.20) & 3.29 (0.13) & 3.01 (0.01) \\ \cline{2-5} 
                         & PI         & \multicolumn{3}{c}{0.93 (0.03)}                                                                 \\ \hline
\multirow{3}{*}{Case 2)} & Efficiency  & \textbf{0.77} (0.02) & \textbf{0.76} (0.02) & 0.56 (0.02) \\ \cline{2-5} 
                         & Dimension         & 9.34 (0.25) & 9.29 (0.26) & 5.39 (0.13)  \\ \cline{2-5} 
                         & PI         & \multicolumn{3}{c}{0.13 (0.03)}                                                                 \\ \hline
\multirow{3}{*}{Case 3)} & Efficiency & \textbf{0.71} (0.02) & \textbf{0.67} (0.02) & 0.55 (0.02)  \\ \cline{2-5} 
                         & Size         & 6.99 (0.23) & 6.61 (0.26) & 4.02 (0.10)  \\ \cline{2-5} 
                         & PI         & \multicolumn{3}{c}{0.35 (0.05)}                                                                 \\ \hline
\end{tabular}
\vspace{-0.3cm}
\end{table}

\textbf{Experiments}:
We now revisit Example~\ref{example:a} in Subsection~\ref{subsec:illustration}, and numerically demonstrate the performances of different methods based on $100$ replications and $n=500$.  For each of the three cases, we compute the means and standard errors of the efficiency (defined in (\ref{eq:efficiency})), dimension of the selected model, and PI, and summarize them in Table~\ref{table:AR}.  
It can be seen that in case 1, BIC and BC perform much better than AIC in terms of efficiency, and PI is close to 1. This is expected from theory as we are in a parametric setting. 
In cases 2 and 3 which are (practically) nonparametric, 
BC performs similarly to AIC,  much better than BIC, and PI is closer to zero. 

%

In practice, AIC seems more widely used compared with BIC,  perhaps mainly due to the thinking that ``all models are wrong'' and minimax-rate optimality of AIC offers {\coco more robustness in adversarial settings} than BIC. {\CO Nevertheless, the parametric setting is still of vital importance. 
First of all, being consistent in selecting the true model if it is really among the candidates is certainly mathematically appealing, and a nonparametric framework can be a practically parametric framework.
More importantly, when decisions need to be made on the use of certain variables, the concept of consistency that avoids over-selection of variables is practically very important. For instance, if medical researchers need to decide if certain genes should be further studied in costly experiments, the protection of over-fitting of BIC avoids recommending variables that are hard to be justified statistically in a followup study, while AIC may recommend quite a few variables that may have some limited predictive values but their effects are too small to be certain with the limited information in the data for decision making and inference purposes. 

The war between AIC and BIC originates from two fundamentally different goals: 
one to to minimize certain loss  for prediction purpose, 
and the other to select the best model for inference purpose.   
A unified view on reconciling such two different goals wherever possible is a fundamental issue in model selection, and it remains an active line of research. We have witnessed some recent advances in that direction and we expect more discoveries to flourish in the future. 


\section{High-Dimensional Variable Selection}  \label{sec:high}

The methods introduced in Section~\ref{sec:criteria} were designed for small models, where the dimension $d_n$ is often required to be $o(\sqrt{n})$ in technical proofs. 
In this section, we elaborate on high-dimensional regression variable selection, an important type of model selection problems in which $d_n$ can be comparable with or even much larger than $n$. To alleviate the difficulties, the data-generating model is often assumed to be {\coco a well-specified linear model, i.e. one of the following candidate models.

Each model $\mathcal{M}$ assumes that 
$y = \sum_{i \in \mathcal{M}} \beta_i x_i + \v$ with $\v$ being random noises. 
Here, with a slight abuse of notation we have also used $\mathcal{M}$ to denote a subset of $\{1,\ldots,d_n\}$, and each data point is written as $z = [y , x_1,\ldots, x_{d_n}]$ with $y$ being the observed response and $x_i$'s being the (either fixed or random) covariates. 
Here, $d_n$ instead of $d$ is used to highlight that the number of candidate variables may depend on the sample size $n$. }

The variable selection problem is also known as support recovery or feature selection in different literature.
The mainstream idea to select the subset of variables is to either solve a penalized regression problem or iteratively pick up  significant variables. The proposed methods differ from each other in terms of how they incorporate unique domain knowledge (e.g. sparsity, multicollinearity, group behavior), or what desired properties (e.g. consistency in coefficient estimation, consistency in variable selection) to achieve. 
{\coco The list of methods we will introduce is far from complete. For example, wavelet shrinkage, iterative thresholding, Dantzig selector, extended BIC, $\ell_q$-regularization with $q \in (0,1)$ (see, e.g.,~\cite{donoho1998minimax,daubechies2004iterative,candes2007dantzig,chen2008extended,foucart2009sparsest}) will not be covered.}

\subsection{Penalized regression for variable selection} 

In a classical setting, a model class is first prescribed to data analysts (either from scientific reasoning or from exhaustive search over $d_n$ candidate variables), and then a criterion is used to select the final model (by applying any properly chosen method explained in Section~\ref{sec:criteria}).  
When there is no ordering of variables known in advance,
and the number of variables $d_n$ is small, {\coco one may simply search over $2^{d_n}$ possible subsets and perform model selection.}
But it is usually computationally prohibitive to enumerate all possible subsets for large $d_n$, especially when $d_n$ is comparable with or even larger than the sample size $n$. 
{\coco Note also that the problem of obtaining a sparse representation of signal $y$ through some chosen basis $x_i$'s (say polynomial, spline, or wavelet basis) usually falls under the framework of variable subset selection as well (but with a different motivation). }
Such a representation can be practically useful in, for example, compressing image signals, locating radar sources, or understanding principal components.

\begin{figure}
\centering
  \includegraphics[width=0.9\linewidth]{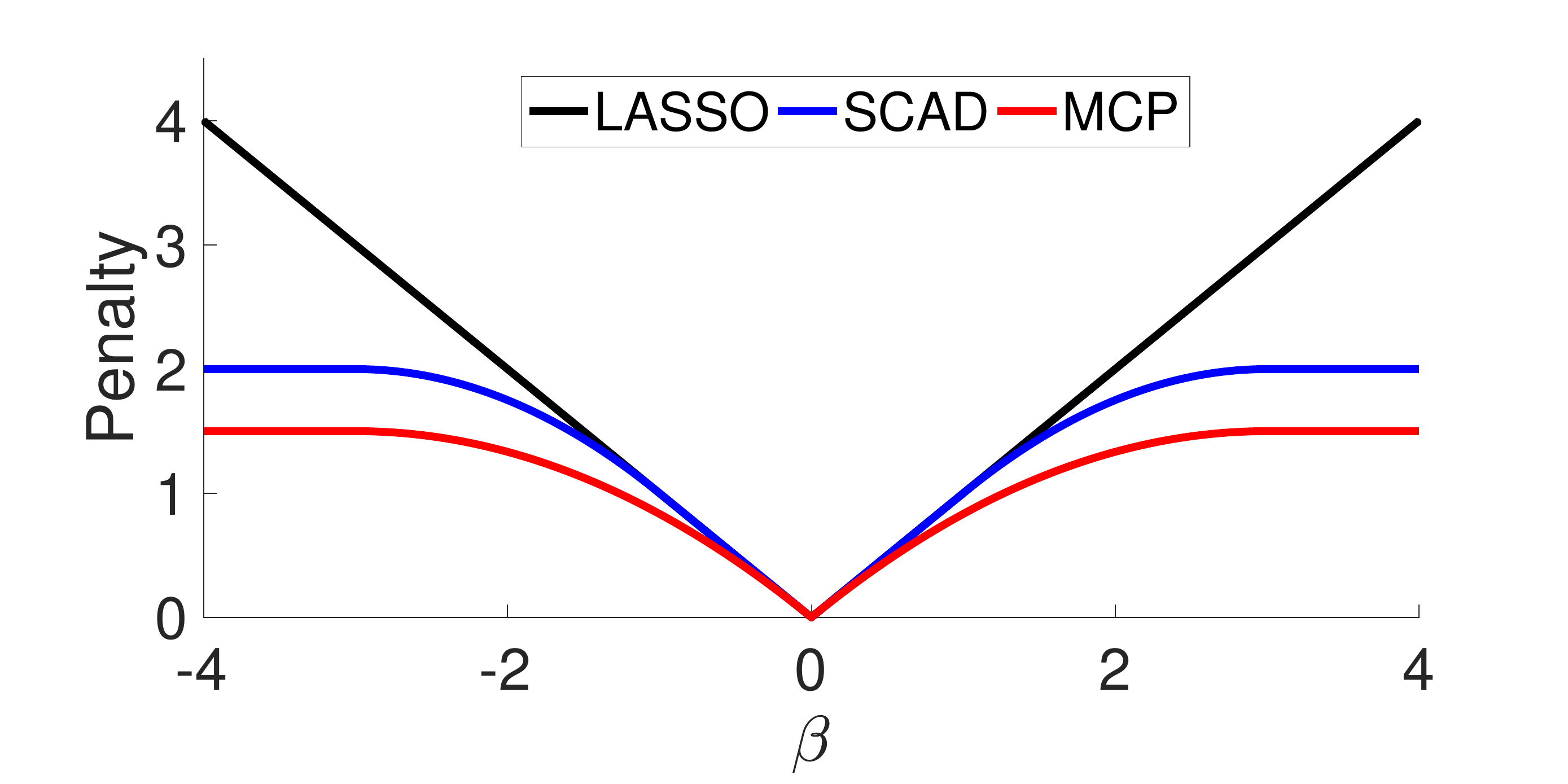}
  \vspace{-0.0in}
  \caption{Illustration of the penalties in LASSO, SCAD, and MCP}
  \label{fig:penalty}
  \vspace{-0.2in}
\end{figure}

Suppose that we have response $ Y_n$ and design matrix $ X_n $ whose entries are $n$ observations of $[y , x_1,\ldots, x_{d_n}]$.
For high-dimensional regression, a popular solution is to consider the following \textit{penalized regression} that amalgamates variable selection and prediction simultaneously in operation. Solve
\begin{align}
\hat{ \beta} = \argmin_{ \beta} \biggl\{ \norm{ Y_n - X_n  \beta}_2^2 + \sum_{j=1}^{d_n} p(|\beta_j|; \lambda, \gamma) \biggr\}  \label{eq:lasso}
\end{align}
and let $\{i: \hat{ \beta}_i \neq 0\}$ be the selected subset of variables.
Here, the $p(\beta ; \lambda,\gamma )$ is a penalty function of $\beta$ with tuning parameters $\lambda, \gamma$ (which are usually determined by cross-validation). It is crucial that the penalty function is not differentiable at $\beta =0$ so that the resulting solution becomes sparse when $\lambda$ gets large.
 
\textbf{Least absolute shrinkage and selection operator} (LASSO)~\cite{tibshirani1996regression} in the form of
$p(\beta ; \lambda ) = \lambda |t|$
is perhaps the most commonly used penalty function. 
Here, $\lambda$ is a tuning parameter which controls the strength of the penalty term. 
Increasing $\lambda$ leads to fewer variables selected. In practice, data analysts can either 1) numerically sweep over a range of $\lambda$, or 2) use the least-angle regression method~\cite{efron2004least} in order to find all the possible candidate models (also called the the solution paths), and then select the model with the best cross-validation performance. {\coco In a time series setting where LASSO solutions need to be continuously updated, fast online algorithms have been proposed (e.g. in~\cite{babadi2010sparls}).}
Given that the data are truly generated by a {\coco linear} model,  tight prediction error bounds have  been established for LASSO. 
Though originally designed for linear regression, LASSO has been also extended to a wide range of statistical models such as generalized linear models (see~\cite{hastie2015statistical} and the references therein). 

\textbf{Smoothly clipped absolute deviation} (SCAD)~\cite{fan2001variable} is another penalized regression that can correct the bias in LASSO estimates that comes from the $\ell_1$-penalty function being unbounded. It was also shown to exhibit ``oracle property'', meaning that as the sample size and model dimension go to infinity, all and only the true variables will be identified with probability going to one, the estimated parameters converge in probability to the true parameters, and the usual asymptotic normality holds as if all the irrelevant variables have already been excluded. 
More discussions on such an oracle property will be included in Section~\ref{sec:clarify}. 
The penalty of SCAD is in the form of 
\begin{align*}
	p(\beta ; \lambda, \gamma ) &= \left\{
	\begin{aligned}
		&\lambda |t| & \textrm{ if } |t|\leq \lambda \\
		& \frac{2 \gamma \lambda |t| - t^2 - \lambda^2}{2 (\gamma - 1)} & \textrm{ if } \lambda <|t|\leq \gamma \lambda \\
		& \frac{ \lambda^2 (\gamma+1)}{2} & \textrm{ if }  |t| > \gamma \lambda  
	\end{aligned}\right. .
\end{align*}
In choosing a parsimonious set of variables, LASSO tends to over-shrink the retained variables. In the SCAD penalty, the idea is to let $\lambda$ and $\gamma$ jointly control that the penalty first suppress insignificant variables as LASSO does, and then tapers off in order to achieve bias reduction. The tuning parameters in SCAD can be chosen by sweeping over a range of them and then applying cross-validation. 

\textbf{Minimax concave penalty} (MCP)~\cite{zhang2010nearly} in the form of 
\begin{align*}
	p(\beta ; \lambda,\gamma ) &= \left\{
	\begin{aligned}
		&\lambda |t| - \frac{t^2}{2 \gamma} & \textrm{ if } |t|\leq \gamma \lambda \\
		& \frac{ \gamma \lambda^2 }{2 } & \textrm{ if } |t| > \gamma \lambda 
	\end{aligned}\right. 
\end{align*}
is a penalized regression that works in a similar way as SCAD. Under some conditions MCP attains minimax convergence rates in probability for the estimation of regression coefficients. 
Fig.~\ref{fig:penalty} illustrates the penalties in LASSO, SCAD, and MCP for $\lambda = 1$ and $\gamma = 3$.

\textbf{Elastic net}~\cite{zou2005regularization} in the form of 
$p(\beta ; \lambda ) = \lambda_1 |t|+ \lambda_2 t^2 $
is proposed to address several shortcomings of LASSO when the covariates are highly correlated. 
The solution $\hat{ \beta}$ of the elastic net penalty exhibits a mixed effects of the LASSO and Ridge penalties. 
Recall that Ridge regression in the form of $p(\beta ; \lambda ) = \lambda t^2$ introduces bias to the regression estimates in order to reduce the large variances of ordinary least squares estimates in the case of multicollinearity, and that LASSO tends to select a sparse subset.
Interestingly, under Elastic net, highly correlated covariates will tend to have similar regression coefficients.
This property, distinct from LASSO, is appealing in many applications when data analysts would like to find all the associated covariates rather than selecting only one from each set of strongly correlated covariates. 

\textbf{Group LASSO}~\cite{yuan2006model} is another penalty  introduced to restrict that all the members of each predefined group of covariates are selected together. Different from (\ref{eq:lasso}), the penalty of the regression is not a sum of $n$ terms, but is replaced with 
$\lambda \sum_{j=1}^{r} \norm{ \beta_{I_j}}_2$
where $ \beta_{I_j}$ is a subvector of $ \beta$ indexed by $I_j$ (the $j$-th group), and $I_1, \ldots, I_r$ form a partition of $\{1,\ldots, n\}$. 
It can be proved that $\hat{ \beta}_{I_j}$ is restricted to be vanishing together for each $j$~\cite{yuan2006model}. The groups are often predefined using prior knowledge. 

\textbf{Adaptive LASSO}~\cite{zou2006adaptive} has been introduced to fit models sparser than LASSO. 
It replaces the penalty in (\ref{eq:lasso}) with $\lambda \sum_{j=1}^{\p} |\tilde{ \beta}_j|^{-u} | \beta_j| $, where $\tilde{ \beta}_j$ is referred to as pilot estimate that can be obtained in various ways (e.g. by least squares for $\p < n$ or univariate regressions for $\p \geq n$). Adaptive LASSO was shown to exhibit the aforementioned oracle property.
The adaptive LASSO can be solved by the same efficient algorithm for solving the LASSO, and it can be easily extended for generalized linear models as well. 

In addition to the above penalized regression, a class of alternative solutions are \textit{greedy algorithms} (or stepwise algorithms), which select a set of variables by making locally optimal decisions in each iteration.


\textbf{Orthogonal matching pursuit} (OMP)~\cite{chen1989orthogonal,pati1993orthogonal}, also referred to as the forward stepwise regression algorithm, is a very popular greedy algorithm which also inspired many other greedy algorithms. 
The general idea of OMP is to iteratively build a set of variables which are ``the most relevant'' to the response. It works in the following way. In each iteration, the variable most correlated with the current residual (in absolute value) is added to the subset (which is initialized as the empty set). Here, the residual represents the component of the observation vector $ y$  not in the linear span of the selected variables.  
Stopping criteria that guarantee good asymptotic properties such as consistency in variable selection remain an active line of research. 
The OMP algorithm can sequentially identify all the significant variables with high probability under some conditions such as weak dependences of the candidate variables (see, e.g.,~\cite{ing2011stepwise,ding2013perturbation} and the references therein).





\textbf{Least-angle regression} (LARS)~\cite{efron2004least} is a greedy algorithm for stepwise variable selection. It can also be used for computing the solution paths of LASSO.  Different from OMP, it doesn't permanently maintain a variable once it is selected into the model. Instead, it only adjusts the coefficient of the most correlated variable until that variable is no longer the most correlated with the recent residual. 
Briefly speaking, LARS works in the following way. It starts with all coefficients $\beta_i$'s being zeros.
In each iteration, {\coco it looks for} the variable $ x_i$ most correlated with the current residual $ r$, and increases its coefficient $\beta_i$ in the direction of the sign of its correlation with $ y$. Once some other variable $ x_j$ has the same correlation with $r$ as $ x_i$ has, {\coco it increases} $\beta_i$ and $\beta_j$ in the direction of their joint least squares until another variable has the same correlation with the residual. {\coco The procedure is repeated until all the variables are in the model or the residuals have become zero. }
%

\subsection{Properties of the penalized regression methods} \label{sec:highDimProperty}

Theoretical examinations of the penalized regression methods have mainly focused on the properties of tight prediction error bounds and consistency in selection. These asymptotic properties are mostly studied by assuming a parametric framework, namely data are truly generated by a linear regression model. Analysis for nonparametric high-dimensional regression models have been also investigated in terms of oracle inequalities for prediction loss~\cite{bickel2009simultaneous} and nonlinear additive models~\cite{lafferty2008rodeo,han2017slants}. 

The goal for prediction in high-dimensional regression focuses the control of the prediction loss (usually squared loss) bound, so that it eventually vanishes even for a very large number of variables $d_n$ (compared with the sample size $n$). 
For instance, suppose that data are generated by 
$
Y_n = X_n  \beta_* +  \v 
$
where $ Y_n \in \mathbb{R}^n,  \beta_* \in \mathbb{R}^{d_n}$, $ \v\sim \mathcal{N}( 0,\sigma^2 I_n)$. 
Let $\norm{\beta_*}_0$ denote the number of nonzero entries in $\beta_*$. 
Then under certain restricted eigenvalue assumptions~\cite{bickel2009simultaneous}, there exist some constants $c_1>2\sqrt{2}$ and $c_2>0$ such that the LASSO solution satisfies $n^{-1}\norm{X_n  \beta_* - X_n \hat{ \beta}}_2^2 \leq  c_2 \sigma^2 \norm{\beta_*}_0 n^{-1} \log d_n $ with probability at least $1-d_n^{1-c_1^2/8}$, if we choose $\lambda = c_1 \sigma \sqrt{n \log \p}$. Note that the above choice of $\lambda$ depends on an unknown $c_1 \sigma$ which, though does not scale with $n$, can have an effect for small sample size. 
Notably, the number of variables $d_n$ is allowed to be much larger than $n$ to admit a good predictive performance, as long as $\log d_n$ is small compared with $n$.   
Similar tight bounds can be obtained by making other assumptions on $\beta_*$ and $X_n$.

Selection consistency, as before, targets the goal of identifying the significant variables for scientific interpretation. The property of asymptotic efficiency we introduced before is rarely considered in high-dimensional regressions, because it is implied by selection consistency in the parametric setting.
For any vector $ \beta \in \mathbb{R}^{d_n}$, let $ r( \beta)$ denote the indicator vector of $ \beta$ such that for any $j=1,\ldots,d_n$, $r_i( \beta)=0$ if $\beta_i=0$, and $r_i( \beta)=1$ otherwise. 
Selection consistency requires that {\coco the probability of $r(\hat{ \beta}) =  r( \beta)$ converges in probability to one (as $n \rightarrow \infty$).} 
Under various conditions such as fixed design or random design matrices, consistency of LASSO in estimating the significant variables has been widely studied under various technical conditions such as sparsity, restricted isometry~\cite{candes2005decoding}, mutual coherence~\cite{donoho2001uncertainty}, irrepresentable condition~\cite{zhao2006model}, and restricted eigenvalue~\cite{bickel2009simultaneous}, which create theoretical possibilities to distinguish the true subset of variables from all the remaining subsets for large $n$. 

At the same time, it has been known that LASSO is not generally consistent in parameter/coefficient estimation. This motivates the methods such as SCAD, MCP, Adaptive LASSO, etc. that correct the estimation bias of LASSO. These three methods are also known to enjoy the so-called \textit{oracle property}.
The oracle property is perhaps more widely considered than selection consistency for high-dimensional regression analysis, since the penalized regression methods target simultaneous parameter estimation and prediction loss control.    
An \textit{oracle estimator}~\cite{fan2001variable} must be consistent in variable selection and parameter estimation, and satisfy 1) the sparsity condition, meaning that 
$
\P\{  r(\hat{ \beta}) =  r( \beta)\} \rightarrow 1
$
as $n \rightarrow \infty$, where the inequality is componentwise; and 2) the asymptotic normality  
$
	\sqrt{n} ( \hat{ \beta}_S -  \beta_S) \limd \mathcal{N}( 0, I^{-1}( \beta_S))
$
where $S$ is the support set of $\beta$, 
$ \beta_S$ is the subvector of $ \beta_*$ indexed by $S$, and $I( \beta_S)$ is the Fisher information knowing $S$ in advance.    
Intuitively speaking, an oracle estimator enjoys the properties achieved by the MLE knowing the true support. We will revisit the oracle property in Subsection~\ref{subsec:oracle}.

\subsection{
Practical 
Performance of penalized regression methods}

With the huge influx of high-dimensional regression data, the penalized regression methods have been widely applied for sparse regression where a relatively small (or tiny) number of variables are selected out of a large number of candidates. For instance, in applications with gene expression type of data, although the number of subjects may be only tens or hundreds, a sparse set of genes is typically selected out of thousand of choices. This has created a lot of excitement, with thousands of publications of such research and applications.  This celebrated sparsity feature of penalized regression methods has generated an optimistic view that even with e.g., fewer than a hundred observations, the modern variable selection tool can identify a sparse subset out of thousands or even many more variables as the set of the most important ones for the regression problem. The estimated model is often readily used for data-driven discoveries.    

There is little doubt that penalized regression methods have produced many successful results for the goal of prediction (see for example \cite{yang2015accurate}). As long as a proper cross-validation is done for tuning parameter selection, the methods can often yield good predictive performances. This said, given the challenge of high-dimension and diverse data sources, the different penalized regression methods may have drastically different relative performances for various data sets. Therefore, a proper choice of a method is important, to which end cross-validation may be used, as will be presented in the next section. 

For the goal of model selection for inference, however, the picture is much less promising.     
Indeed, many real applications strongly suggest that the practice of using the selected model for understanding and inference may be far from being reliable. It has been reported that the selected variables from  these penalized regression methods are often severely unstable, in the sense that the selection results can be drastically different under a tiny perturbation of data (see~\cite{nan2014variable} and the references therein). Such high uncertainty damages reproducibility of the statistical findings~\cite{ioannidis2005most}.
Overall, being overly {\coco optimistic} about the interpretability of high-dimensional regression methods can lead to spurious scientific discoveries.  

The fundamental issue still lies in the potential discrepancy between inference and prediction, which is also elaborated in Section~\ref{sec:AICBIC} and Subsection~\ref{subsec:oracle}. If data analysts know in advance that the true model is exactly (or close to) a stable low-dimensional linear model, then the high-dimensional methods with the aforementioned oracle property may produce stable selection results not only good for prediction but also for inference purposes. Otherwise, the produced selection is so unstable that analysts can only focus on prediction alone.  
In practice, data analysts may need to utilize data-driven tools such as model averaging~\cite{yang2001adaptive}, resampling~\cite{meinshausen2010stability}, confidence set for models~\cite{ferrari2015confidence}, or model selection diagnostic tools such as the parametricness index introduced in Section~\ref{sec:AICBIC} in order to make sure the selected variables are stable and properly interpretable. 
Considerations along these lines also lead to stabilized variable selection methods~\cite{meinshausen2010stability,lim2016estimation,yang2017toward}.
The instability of penalized regression also motivated some recent research on 
post-model-selection inference~\cite{berk2013valid,taylor2014post}. Their interesting results in specific settings call for more research for more general applications.  

\section{Modeling Procedure Selection} \label
{sec:selecting}

The discussions in the previous sections have focused on \textit{model selection} in the narrow sense where the {\coco candidates are models.} In this section, we review the use of CV as a general tool for \textit{modeling procedure selection}, which aims to select one from a finite set of modeling procedures~\cite{zhang2015cross}.  
For example, one may first apply modeling procedures such as AIC, BIC, and CV for variable selection to the same data, and then select one of those procedures (together with the model selected by the procedure), using an appropriately designed CV (which is at the second level). 
Another example is the emerging online competition platforms such as Kaggle, that compare new problem-solving procedures and award prizes using cross-validation.
{\coco The ``best'' procedure is defined in the sense that it outperforms, with high probability, the other procedures in terms of out-sample prediction loss for sufficiently large $n$ (see for example \cite[Definition 1]{yang2006comparing}). }

There are two main goals of modeling procedure selection.
The first is to \textit{identify} with high probability the best procedure among the candidates. The property of selection consistency is of interest here. 
The second goal of modeling procedure selection is to approach the {\coco best performance (in terms of out-sample prediction loss)} offered by the candidates, instead of pinpointing which candidate procedure is the best. Note again that in case there are procedures that have similar best performances, we do not need to single out the best candidate to achieve the asymptotically optimal performance.  

Similarly to model selection, for the task of modeling procedure selection, CV randomly splits $n$ data into $n_t$ training data and $n_v$ validation data (so $n=n_t+n_v$). The first $n_t$ data are used to run different modeling procedures, and the remaining $n_v$ data are used to select the better or best procedure.
We will see that for the first goal above, the evaluation portion of CV should be large enough; while for the second goal, a smaller portion of the evaluation may be enough to achieve optimal predictive performance.


In the literature, much attention has been focused on choosing whether to use the AIC procedure or BIC procedure for data analysis. 
For regression variable selection, it has been proved that the CV method is consistent in choosing between AIC and BIC given $n_t \rightarrow \infty$, $n_v/n_t \rightarrow \infty$, and some other regularity assumptions~\cite[Thm. 1]{zhang2015cross}. In other words, the probability of BIC being selected goes to $1$ in a parametric framework, and the probability of AIC being selected goes to $1$ otherwise. In this way, the modeling procedure selection using CV naturally leads to a hybrid model selection criterion that builds upon strengths of AIC and BIC. Such a hybrid selection combines some theoretical advantages of both AIC and BIC. This aspect is to be clearly seen in the context of Section~\ref{sec:AICBIC}.
The task of classification is somewhat more relaxed compared with the task of regression. In order to achieve consistency in selecting the better  classifier, the splitting ratio may be allowed to converge to infinity or any positive constant, depending on the situation~\cite{yang2006comparing}. In general, it is safe to let $n_t \rightarrow \infty$ and $n_v/n_t \rightarrow \infty$ for consistency in modeling procedure selection. 

Closely related to the above discussion is the following paradox. 
Suppose that a set of newly available data is given to an analyst. The analyst would naturally add some of the new data in the training phase and some in the validation phase. Clearly, with more data added to the training set,  each candidate modeling procedure is improved in accuracy; with more data added to the validation set, the evaluation is also more reliable. 
It is tempting to think that improving the accuracy on both training and validation would lead to sharper comparison between procedures.
However, this is not the case. The prediction error estimation and procedure comparison are two different targets. 

\textbf{Cross-validation paradox}: Better training and better estimation (e.g., in both bias and variance) of the prediction error by CV together do {\it not} imply better modeling procedure selection~\cite{zhang2015cross}.
Intuitively speaking, when comparing two procedures that are naturally close to each other, the improved estimation accuracy by adopting more observations in the training part only makes the procedures more difficult to be distinguishable. The consistency in identifying the better procedure cannot be achieved unless the validation size diverges fast enough.   

\begin{table}[tb]
\centering
\caption{Cross-validation paradox: More observations in training and evaluations do not lead to higher selection accuracy in selecting the better procedure.}
\label{table:syntheticCV}
\begin{tabular}{cccccc}
\hline
sample size $n$    & 100    & 200     & 300     & 400     & 500    \\ \hline
Training size $n_t$    & 20    & 70     & 120     & 170     & 220    \\ \hline
Accuracy & 98.3\% & 94.9\% & 93.7\% & 92.3\% & 92.5\%     \\ \hline
\end{tabular}
\end{table}

\textbf{Experiments}:
We illustrate the cross-validation paradox using the synthetic data generated from the linear regression model $y = \beta_1 x_1 + \beta_2 x_2 + \beta_3 x_3 + \v$, where $\beta = [1,2,0]^\T$, the covariates $X_j$ ($j=1,2,3$) and noise $\v$ are independent standard Gaussian. Given $n$ observations $(y_i,x_{1,i},x_{2,i},x_{3,i})_{i=1,\ldots,n}$, we compare the following two different uses of linear regression. The first is based on $X_1$ and $X_2$, and the second is based on all the $3$ covariates. Note that in this experiment, selecting the better procedure is equivalent to selecting a better model.  
The data-generating model indicates that $x_3$ is irrelevant for predicting $y$, so that the first procedure should be better than the second. Suppose that we start with $100$ observations. We randomly split the data $100$ times, each with $20$ training data and $80$ validation data, and record which procedure gives the smaller average quadratic loss during validation. We then add $50$ new data to the training set and $50$ to the validation set, and record again which procedure is favored. We continuing doing this until the sample size reaches $500$. By running $1000$ independent replications, we summarize the frequency of the first procedure being favored in Table~\ref{table:syntheticCV}. As the paradox suggests, the accuracy of identifying the better procedure does not necessarily increase when more observations are added to both the estimation phase and the validation phase.

\section{Clarification of Some Misconceptions} \label{sec:clarify}


\subsection{Pitfall of one-size-fits-all recommendation of data splitting ratio of cross-validation}

There are wide-spread general recommendations on how to apply cross-validation for model selection. For instance, it is common to use 10-fold CV for model selection. Such guidelines seem to be unwarranted. First, it mistakenly disregards the goal of model selection. For prediction purposes, leave-one-out is actually preferred in tuning parameter selection for traditional nonparametric regression. In contrast, for selection consistency, 10-fold often leaves too few observations in evaluation to be stable. Indeed, 5-fold often produces more stable selection results for high-dimensional regression. {\coco Second, $k$-fold CV, regardless of $k$,  in general, is often unstable in the sense that a  different dividing of data can produce a very different selection result. A common way to improve performance is to randomly divide the data into $k$ folds several times and use the average validation loss for selection.}

For \textit{model selection}, CV randomly splits $n$ data into $n_t$ training data and $n_v$ validation data. 
Common practices using 5-fold, 10-fold, or 30\%-for-validation do not exhibit asymptotic optimality (neither consistency nor asymptotic efficiency) in simple regression models, and their performances can be very different depending on the goal of applying CV. 
In fact, it is known that the delete-$n_v$ CV is asymptotically equivalent to GIC$_{\lambda_n}$ with 
$\lambda_n = n / (n-n_v) + 1 $
under some assumptions~\cite{shao1997asymptotic}.
It is also known that GIC$_{\lambda_n}$ achieves asymptotic efficiency in a nonparametric framework only with $\lambda_n=2$, and asymptotic efficiency in a parametric framework only with $\lambda_n \rightarrow \infty$ (as $n \rightarrow \infty$). 
In this context, from a theoretical perspective, the optimal splitting ratio $n_v/n_t$ of CV should either converge to zero or diverge to infinity in order to achieve asymptotic efficiency, depending on whether the setting is nonparametric or parametric. 


For \textit{modeling procedure selection}, 
it is often necessary to let the validation size take a large proportion (e.g., half) in order to achieve good selection accuracy. 
In particular, the use of LOO for the goal of comparing procedures is the least trustworthy (see Section~\ref{sec:selecting}).  


\begin{table}[tb]
\centering
\caption{Classification for handwritten digits: smaller $n_v/n_t$ tends\ to give better predictive performance}
\label{table:realCV}
\begin{tabular}{cccccc}
\hline
Ratio   & 0.95    & 0.9     & 0.5     & 0.1     & 0.05    \\ \hline
Accuracy & 72.24\% & 90.28\% & 91.47\% & 91.47\% & 92.99\%     \\ \hline
\end{tabular}
\vspace{-0.4cm}
\end{table}

\textbf{Experiment}:
We show how the splitting ratio can affect CV for model selection using the Modified National Institute of Standards and Technology (MNIST) database~\cite{lecun1998gradient}, which consists of $70,000$ images of handwritten digits (from 0-9) with $28 \times 28$ pixels. 
We implement $6$ candidate feed-forward neural network models for classification. The first $4$ models have $1$ hidden layer, and the number of hidden nodes are respectively $17,18,19,20$; the $5$th model has $2$ hidden layers with $20$ and $4$ nodes; the $6$th model has $3$ hidden layers with $20$, $2$, and $2$ nodes.
Since the true data-generating model for the real data is unavailable, we take 35,000 data (often referred to as the test data) out for approximating the true prediction loss,
	and use the remaining data to train and validate.
For model selection, we run CV with different  $n_v/n_t$. For each ratio, we  compute the average validation loss of each candidate model based on 10 random partitions. We then select the model with the smallest average loss, and calculate its ``true'' predictive loss using the remaining 35,000 data. 
The results recorded in Table~\ref{table:realCV} indicate that a smaller splitting ratio $n_v/n_t$ leads to better classification accuracy. 
This is in line with the existing theory, since neural network models are likely to be ``nonparametric". 
This example also provides a complementing message to the cross-validation paradox. At ratio $0.95$, the training sample size is too small to represent the full sample size, so the ranking of the candidate models estimated from training data can be unstable and deviate from the ranking of models estimated from the full dataset. 

\subsection{Since all models are wrong, why pursing consistency in selection?} 
\label{subsec:consistency}


Since the reality is usually more complicated than a parametric model, perhaps everyone agrees that all models are wrong and that the consistency concept of selecting the true model\footnote{As mentioned before, typically the true data-generating model is the best model in a parametric framework.} in a parametric framework is irrelevant. 
{One view on such selection consistency} is that in many situations, a stable parametric model can be identified and it can be treated as the ``true model". 
Such an idealization for theoretical investigation with practical implications is no more sinful than deriving theories under nonparametric assumptions. The true judge should be the performance in real applications.  
On the other hand, the notion of ``consistency'' in a nonparametric  framework, however, is rarely used in the literature. In fact, it was shown that there does not exist any model selection method that can guarantee consistency in nonparametric regression settings (see, e.g.,~\cite{DingBridge2}). This partly explains why the concept of asymptotic efficiency (which is a weaker requirement) is more widely used in nonparametric frameworks.

\subsection{Controversy over the oracle property}
\label{subsec:oracle}

The popular {\it oracle property} (as mentioned in Subsection~\ref{sec:highDimProperty}) for high-dimensional variable selection has been a focus in many research publications. However, it has been criticized by some researchers (see, e.g.,~\cite{leeb2008sparse}). 
At first glance, the oracle property may look very stringent. But we note that its requirement is fundamentally only as stringent as consistency in variable selection. {\coco In fact, if all the true variables can be selected with probability tending to one by any method, then one can obtain MLE or the like restricted to the relevant variables for optimal estimation of the unknown parameters in the model.} 
To our knowledge, there is  neither claim nor reason to believe that the original estimator should be better than the re-fitted one by MLE based on the selected model. 
Though the oracle property is not theoretically surprising beyond consistency, it is still interesting and nontrivial to obtain such a property with only one stage of regression (as SCAD, MCP, and Adaptive LASSO do). These methods, when armed with efficient algorithms, may save the computational cost in practice.   

 

It was emphasized in \cite{leeb2008sparse} that the oracle estimator does not perform well in a uniform sense for point or interval estimation of the parameters.
A paid price for the oracle property is that the risk of any ``oracle estimator'' (see\cite{fan2001variable}) 
has a supremum that diverges to infinity, {\coco i.e., 
$$ 
\sup_{ \beta \in \mathbb{R}^p} E_{\beta}\{ n (\hat{\beta} - \beta)^\T (\hat{\beta} - \beta) \} \rightarrow \infty
$$
as sample size $n \rightarrow \infty$
(see for instance~\cite{leeb2008sparse}). Here, we let $E_{\beta}$ denote expectation with respect to the true linear model with coefficients $\beta$.} 
In fact, for any consistent model selection method, we can always find a parameter value that is small enough so that the selection method tends to not include it (since it has to avoid over-selection), yet the parameter value is big enough so that dropping it has a detrimental effect in rate of convergence (see, e.g.,~\cite{yang2005can, yang2007prediction}). 
While uniformity and robustness are valid and important considerations, we do not need to overly emphasize such properties. Otherwise we are unduly burdened to retain not very useful variables in the final model and have to lose the ability in choosing a practically satisfying parsimonious model for interpretation and inference. 

\section{Some general recommendations}
\label{recomm}

Model selection, no matter how it is done, is exploratory in nature and cannot be confirmatory. Confirmatory conclusions can only be drawn based on well-designed followup studies. Nevertheless, good model selection tools can provide valuable and reliable information regarding explanation and prediction.       
Obviously there are many specific aspects of the data, nature of the models and practical considerations of the variables in the models, etc., that make each model selection problem unique to some degree. Nonetheless, based on the literature and our own experiences, we give some general recommendations.

\begin{enumerate}

\item Keep in mind the main objective of model selection. 

a) If one needs to declare a model for inference, model selection consistency is the right concept to think about. Model selection diagnostic measures need to be used to assess the reliability of the selected model. 
In a high-dimensional setting, penalized regression methods are typically highly uncertain. 
{\coco For selection stability, when choosing a tuning parameter by cross-validation, for instance, 5-fold tends to work better than 10-fold (see, e.g.,~\cite{nan2014variable}). }
 
b) If one's main goal is prediction, model selection instability is less of a concern, and any choice among the best performing models may give a satisfying prediction accuracy. In a parametric framework, consistent selection leads to asymptotic efficiency. 
In a nonparametric framework, selection methods based on the optimal tradeoff between estimation error and approximation error lead to asymptotic efficiency.
When it is not clear if a (practically) parametric framework is suitable, we recommend the use of an adaptively asymptotic efficient method (e.g., the BC criterion). 

\item When model selection is for prediction, the minimax consideration gives more protection in the worst case. 
{\coco If one  postulates that the nature is adversary,} the use of a minimax optimal criterion (e.g., AIC) is safer (than e.g., BIC).   

\item When prediction is the goal, one may consider different types of models and methods and then apply cross-validation to choose one for final prediction. If one needs to know which model or method is really the best, a large enough proportion (e.g., 1/3 or even half) for validation is necessary. If one just cares about the prediction accuracy and has little interest in declaring the chosen one being the best, the demand on the validation size may be much lessened.

\end{enumerate}

\section*{Acknowledgement}
The authors thank 
Dr. Shuguang Cui and eight anonymous reviewers for giving feedback on the initial submission of the manuscript.
The authors are also grateful to Dr. Matthew McKay,
Dr. Osvaldo Simeone, and three anonymous reviewers for handling the full submission of the manuscript. 
We owe the three anonymous reviewers for their comprehensive comments that have greatly improved the paper.

\ifCLASSOPTIONcaptionsoff
  \newpage
\fi


\bibliographystyle{IEEEtran}
\balance
\bibliography{references_com,overview}
%

%
%

\end{document}